\documentclass{article}

\PassOptionsToPackage{numbers,compress}{natbib}
\usepackage[preprint]{neurips_2026}

\usepackage[utf8]{inputenc}
\usepackage[T1]{fontenc}
\usepackage[english]{babel}
\usepackage{amsmath,amssymb,amsthm,bm}
\usepackage{booktabs}
\usepackage{array}
\usepackage{tabularx}
\usepackage{longtable}
\usepackage{graphicx}
\usepackage{tikz}                                   
\usepackage{xcolor}
\usepackage{enumitem}
\usepackage{caption}
\usepackage{microtype}
\usepackage{url}
\usepackage{hyperref}

\newcommand{\dupnote}[1]{}

\usepackage{cleveref}
\usepackage{xspace}

\definecolor{figblue}{HTML}{7EACCE}    
\definecolor{figgreen}{HTML}{7AAC5C}   
\definecolor{figred}{HTML}{BE4C49}     
\definecolor{figorange}{HTML}{D08A3E}  

\hypersetup{colorlinks=true,linkcolor=figblue!65!black,citecolor=figgreen!55!black,
            urlcolor=figblue!65!black,breaklinks=true}

\newcommand{\R}[1]{\mathbb{R}^{#1}}
\newcommand{\M}{\mathcal{M}}
\newcommand{\xv}{x_v}
\newcommand{\xe}{x_e}
\newcommand{\xf}{x_f}
\newcommand{\dzero}{d_0}
\newcommand{\done}{d_1}
\newcommand{\Lzero}{L_0}

\newcommand{\ux}{u_x}
\newcommand{\uy}{u_y}
\newcommand{\nut}{\nu_t}
\newcommand{\Uinf}{U_\infty}
\newcommand{\relL}{\text{rel-}L_2}
\newcommand{\CL}{C_L}
\newcommand{\rhoL}{\rho_L}
\newcommand{\simplex}{\Delta^{K}}
\newcommand{\gate}{g}
\newcommand{\Nv}{N}
\newcommand{\Ne}{N_e}
\newcommand{\Nf}{N_f}

\newcommand{\code}[1]{\texttt{\small #1}}
\newcommand{\best}[1]{\textbf{#1}}

\theoremstyle{plain}

\theoremstyle{remark}

\setlist{itemsep=1pt,topsep=2pt,parsep=0pt}
\let\oldthebibliography\thebibliography
\renewcommand{\thebibliography}[1]{%
  \oldthebibliography{#1}\footnotesize%
  \setlength{\itemsep}{0pt}\setlength{\parsep}{0pt}\setlength{\parskip}{0pt}}

\title{Learnable Hodge--Laplacian Driven Mixture of Experts for Steady-PDE Surrogates on Coarse Unstructured Meshes}

\author{%
  \textbf{Stavros Nousias}$^{1}$, 
  \textbf{Akis Nousias}$^{2}$, 
  \textbf{Konstantinos Gkrispanis}$^{1}$ \\[0.5em]
  $^{1}$Technical University of Munich \\
  $^{2}$K3Y Labs \\
}

\date{}

\begin{document}
\maketitle

\newcommand{\overviewafterabstract}{}
\begin{abstract}
Surrogate models for steady partial differential equations on unstructured meshes often apply a single mixing mechanism to the whole solution. We propose a mixture of experts architecture supervised at different spectral scales. A controller that utilizes learnable laplacian matrices, referred to as Hodge Controller, receives the mesh geometry and input features and produces per-vertex weights that combine the expert feature vectors. During training, a separate discrete Laplacian $\Lzero=\star_0^{-1}\dzero^{\!\top}\star_1\dzero$ decomposes the reference fields into scale-dependent targets for auxiliary expert heads. We evaluate the model on AirFRANS, a steady Reynolds-averaged Navier--Stokes (RANS) airfoil benchmark, using remeshed triangular grids with approximately $9.6$k faces and $4.9$k vertices. On the official $800/200$ split, our model achieves $31\%$ lower relative $L_2$ error and $47\%$ lower interior mean squared error than Transolver on the same evaluation meshes, with approximately half as many parameters. Training allocations and checkpoint-selection procedures differ between our model and the baselines. Code: \url{https://github.com/snousias/hodge-moe-pde}.
\end{abstract}

\ifdefined\overviewafterabstract
  \par\vspace{-1.5ex}\noindent\begin{minipage}{\textwidth}
  \captionsetup{type=figure}
\else
  \begin{figure}[t]
\fi
  \centering
  \ifdefined\overviewafterabstract
  \includegraphics[width=0.88\textwidth]{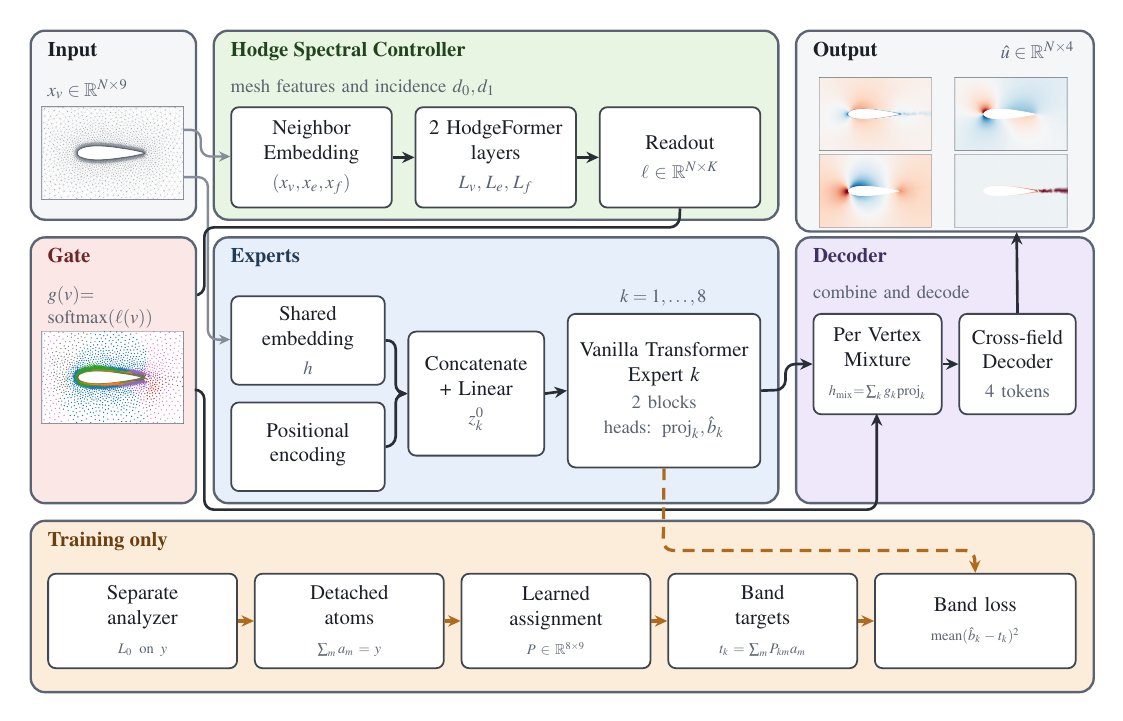}
  \else
  \includegraphics[width=\textwidth]{figures/overview_landscape.pdf}
  \fi
  \caption{Overview of the model. The Hodge controller routes transformer experts that combine shared embeddings with positional encodings at different frequency scales. Their representations are mixed and decoded into four fields. During training, a separate analyzer decomposes the reference fields into atoms, which a learned assignment combines into targets for the auxiliary expert heads (dashed path).}
  \label{fig:arch}
\ifdefined\overviewafterabstract
  \end{minipage}\par\medskip
\else
\end{figure}
\fi

\section{Introduction}
\label{sec:intro}

Computational fluid dynamics is central to aerodynamic design. Solving Reynolds-averaged Navier--Stokes (RANS) equations for steady turbulent flows~\citep{bonnet2022airfrans} on a discretisation of the domain allows engineers to evaluate a candidate geometry's pressure, velocity and turbulence fields. Evaluating many candidates this way is computationally expensive, which limits how many candidate geometries can be evaluated~\citep{brunton2020machine,vinuesa2022enhancing}. 

Learned surrogate models predict solution fields from geometry, boundary data and flow parameters~\citep{thuerey2020deep}. The fields are obtained by evaluating the network once, which is orders of magnitude faster at inference. Physics-informed networks encode the governing equations in the training objective, so the output satisfies the PDE residual and the boundary conditions~\citep{raissi2019physics,karniadakis2021physics}. This approach requires an explicit closure model, which is restrictive for industrial RANS turbulence. Neural operators instead learn the input-output map from simulation data using parameters shared across discretisations~\citep{kovachki2023neural,azizzadenesheli2023neural}. The Fourier neural operator~\citep{li2021fourier} approximates the kernel by a pointwise multiplication in the Fourier domain, and extensions to irregular domains use message passing~\citep{li2020neural}. Mesh deformation~\citep{li2023fourier}, or message passing directly on the mesh~\citep{pfaff2021learning}, local kernels approximate the global coupling of elliptic problems poorly, and Fourier bases lose accuracy on complex geometries. Transformer solvers address this coupling directly, at the cost of a quadratic attention complexity~\citep{cao2021choose,li2022transformer,hao2023gnot,li2023scalable}. Among these, Transolver~\citep{wu2024transolver} assigns mesh points to physics slices and attends among the resulting tokens. The total cost still depends on the number of mesh points. Each of these architectures uses a single type of feature-update operation to mix information.

A separate line of work uses discrete mesh differential operators in the network, using discrete exterior calculus~\citep{desbrun2005discrete}. In HodgeNet ~\citep{smirnov2021hodgenet}, hodge stars are learned as diagonal matrices and in DiffusionNet as learned diffusion times~\citep{sharp2022diffusionnet}. HodgeFormer ~\citep{nousias2026hodgeformer} proposes attention-parametrised stars on all three cochain degrees. In all of these constructions, the learned operator lies on the \emph{prediction} path, computing the field directly. 

Sparsely gated mixtures of experts are normally used to increase the parameter count while evaluating only a subset of experts for each token, with a router implemented as a linear map from token features to routing logits, under which the specialisation the experts acquire is not fixed by the construction~\citep{shazeer2017moe,fedus2022switch}. Geometry-gated mixtures of experts for multi-scale PDE problems are available in the literature. GNOT~\citep{hao2023gnot} gates a mixture of experts with a network that takes query coordinates as input, and GAOT~\citep{wen2025geometry} combines a ladder of scale-specific branches with a learned per-point softmax gate. We adopt this pattern and use a Hodge-Laplacian controller, and propose a mixture of plain transformer experts, each biased towards one spectral scale. We supervise each expert with an auxiliary target constructed from the reference field. A second, diagonal-star Hodge Laplacian decomposes the training target into a sequence of components. 

The contributions of this paper can be summarized as follows: (i) We discuss a Gaussian star construction (\cref{sec:controller}) that is symmetric positive definite whenever its neighbour support is mutual, and under which the assembled operator then has a real non-negative spectrum. (ii) We propose an architecture that uses the learned discrete Hodge Laplacian to compute routing weights. (iii) We introduce a supervision scheme in which a second, diagonal-star learnable Laplacian decomposes the target into a sequence of components (\cref{sec:analyzer}). (iii) We evaluate the approach on the official AirFRANS split against a Transolver baseline, trained on the same working meshes with identical data, loss and metric code (\cref{sec:results}). \Cref{sec:notation} lists every symbol used below, with its type and its meaning. The remainder of this paper is organized as follows: \cref{sec:rw} reviews related work, \cref{sec:method} presents the method, \cref{sec:exp} describes the simulation setup and reports the results, and \cref{sec:discussion} discusses the difference in lift-coefficient error and the limits of the comparison.


\section{Related Work}
\label{sec:rw}

\noindent\textbf{Physics-informed and data-driven surrogates.}
Physics-informed neural networks learn PDE solution fields by incorporating
the governing equations and boundary conditio
training~\citep{raissi2019physics,karniadakis2021physics}.
Physics-informed neural operators extend thisn
operators across families of PDE inputs~\citep{li2024physics}.
For RANS flows, this requires specifying the mean-flow equations and a
turbulence closure. Data-driven surrogates instead train on simulated
solution fields and specify no governing
equation~\citep{thuerey2020deep}. The AirFRANS benchmark provides such
fields for airfoil flows~\citep{bonnet2022airfrans}.

\textbf{Neural operators on irregular domains.}
Neural operators learn maps between function spaces using parameters shared across discretisations~\citep{kovachki2023neural,azizzadenesheli2023neural}. 
DeepONet represents an operator through a branch network that encodes the input function and a trunk network that encodes the query location~\citep{lu2021learning}, whereas the Fourier neural
operator (FNO) uses learned spectral multipliers to implement global convolutions~\citep{li2021fourier}. Graph kernel networks~\citep{li2020neural}, learned coordinate deformations (Geo-FNO)~\citep{li2023fourier}, and graph-to-grid encoding with a latent FNO (GINO)~\citep{li2023geometry} support irregularly sampled inputs and different geometries. 
MeshGraphNets instead learn local interactions along mesh edges~\citep{pfaff2021learning}. For local message passing, exchanging information between distant vertices requires more message-passing layers or additional multiscale mechanisms. Spectral and latent-grid methods provide alternative ways to exchange global information. These choices concern how features propagate, rather than explicitly supervising separate predictors at different mesh spectral scales.

\textbf{Transformer PDE solvers.}
Dense attention enables global interactions but has quadratic cost in the number of mesh points.
Galerkin attention~\citep{cao2021choose}, OFormer~\citep{li2022transformer},
GNOT~\citep{hao2023gnot}, and FactFormer~\citep{li2023scalable} use linear or factorised
attention, while inducing-point and anchored-branched architectures process reduced token
sets~\citep{lee2024inducing,bleeker2025abupt}. Transolver~\citep{wu2024transolver} assigns each point weights over learned groups called physics slices, applies attention among slice tokens, and maps the result back
to the points. 
For a fixed slice count, its cost grows linearly with mesh size. We use Transolver as the principal baseline under a shared remeshing and training protocol. Our dense transformer experts retain quadratic attention cost; the proposed controller addresses expert specialisation rather than removing this scaling limitation.

\textbf{Geometric gating and multiscale mixtures.}
Sparsely gated mixtures of experts increase capacity by activating only a subset of experts per
token~\citep{shazeer2017moe,fedus2022switch}. Geometric and multiscale gating already appear in
PDE operator learning: GNOT mixes feed-forward experts using weights computed from the query coordinates,
interpreting the gate as soft domain decomposition~\citep{hao2023gnot}. 
GAOT combines graph operator branches at different neighbourhood radii through learned softmax weights, together with a latent transformer processor~\citep{wen2025geometry}. Our contribution builds on this precedent by using a mesh Hodge-Laplacian controller to mix full transformer experts at each vertex, and by supervising auxiliary expert heads with mesh spectral bands. Fourier-features directs the expert inputs towards different spatial scales~\citep{tancik2020fourier}; band supervision supplies an additional constraint on their auxiliary predictions.

\textbf{Learned differential operators and spectral analysis.}
Discrete exterior calculus separates mesh topology represented by signed incidence matrices from metric information encoded
in Hodge stars~\citep{desbrun2005discrete}. Hodge Laplacians extend graph differential operators
to higher-degree cochains~\citep{lim2020hodge,schaub2020random}. The seminal work of HodgeNet learns discrete Hodge
stars for spectral geometry processing~\citep{smirnov2021hodgenet}, DiffusionNet learns diffusion
times for surface feature propagation~\citep{sharp2022diffusionnet}, and DiffGCN constructs graph
convolutions from differential operators~\citep{eliasof2020diffgcn}. 
HodgeFormer parameterises stars through local attention to construct operators on vertices, edges, and
faces~\citep{nousias2026hodgeformer}. Related neural PDE models use structure-preserving finite
element formulations~\citep{shaffer2026structure}, Hodge-decomposed solution
spaces~\citep{zheng2026topology}, or cross-dimensional interactions on cell
complexes~\citep{bastian2026topological}. 


\section{Methodology}
\label{sec:method}
\subsection{Overview}
\label{sec:arch}
We propose a mixture of transformer experts for steady-PDE prediction on unstructured meshes, with per-vertex weights computed by a learnable Laplacian controller. The model predicts velocity components $\ux,\uy$, pressure $p$, and turbulent kinematic viscosity $\nut$ at the vertices of an unstructured mesh.  The prediction path consists of a shared embedding, a controller, transformer experts, and a field decoder. The controller produces routing weights from vertex, edge, and face features together with the mesh incidence matrices and neighborhood indices. Each expert combines the shared embedding with fixed Fourier positional features and produces both a latent representation and an auxiliary prediction. The routing weights combine the latent representations, which the decoder maps to the four output fields. During training, a separate diagonal-star vertex Laplacian smooths the reference fields to construct telescoping atoms. A learned assignment combines these atoms into expert-specific supervision targets. The controller uses two \textbf{PSD-Hodge layers}, which keep the block structure of HodgeFormer~\citep{nousias2026hodgeformer} and replace its softmax-parameterized stars with Gaussian stars intended to give a positive-semidefinite (PSD) construction.

\ifdefined\overviewafterabstract\else
\ifdefined\overviewafterabstract
  \par\vspace{-1.5ex}\noindent\begin{minipage}{\textwidth}
  \captionsetup{type=figure}
\else
  \begin{figure}[t]
\fi
  \centering
  \ifdefined\overviewafterabstract
  \includegraphics[width=0.88\textwidth]{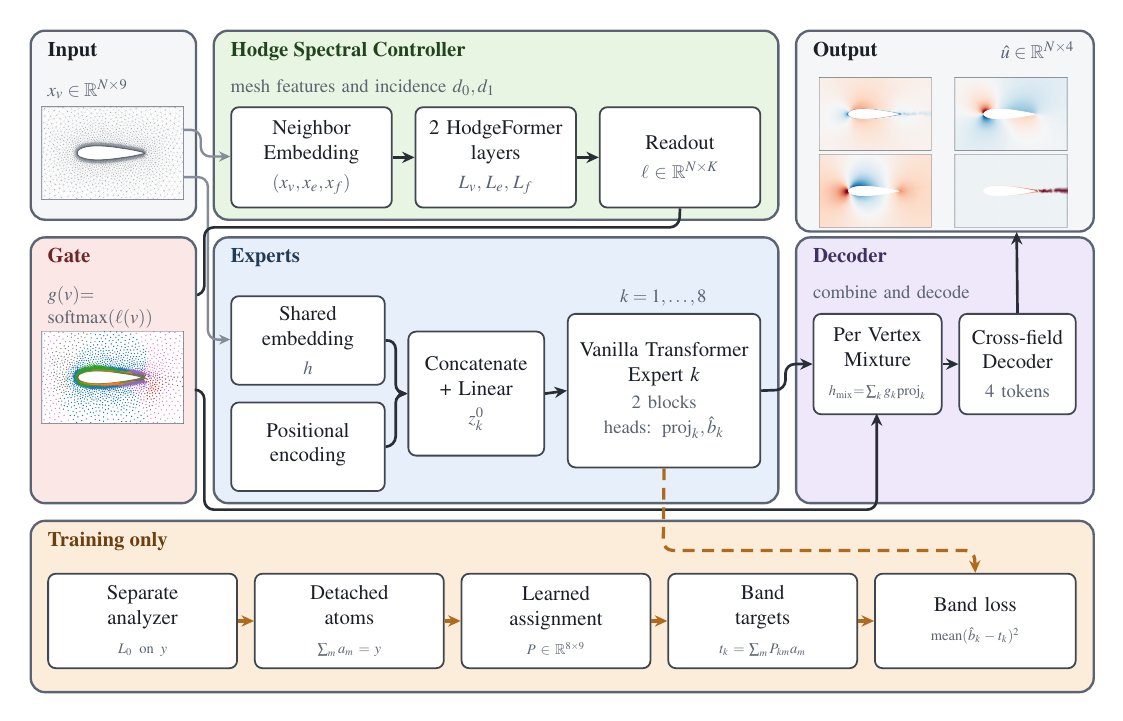}
  \else
  \includegraphics[width=\textwidth]{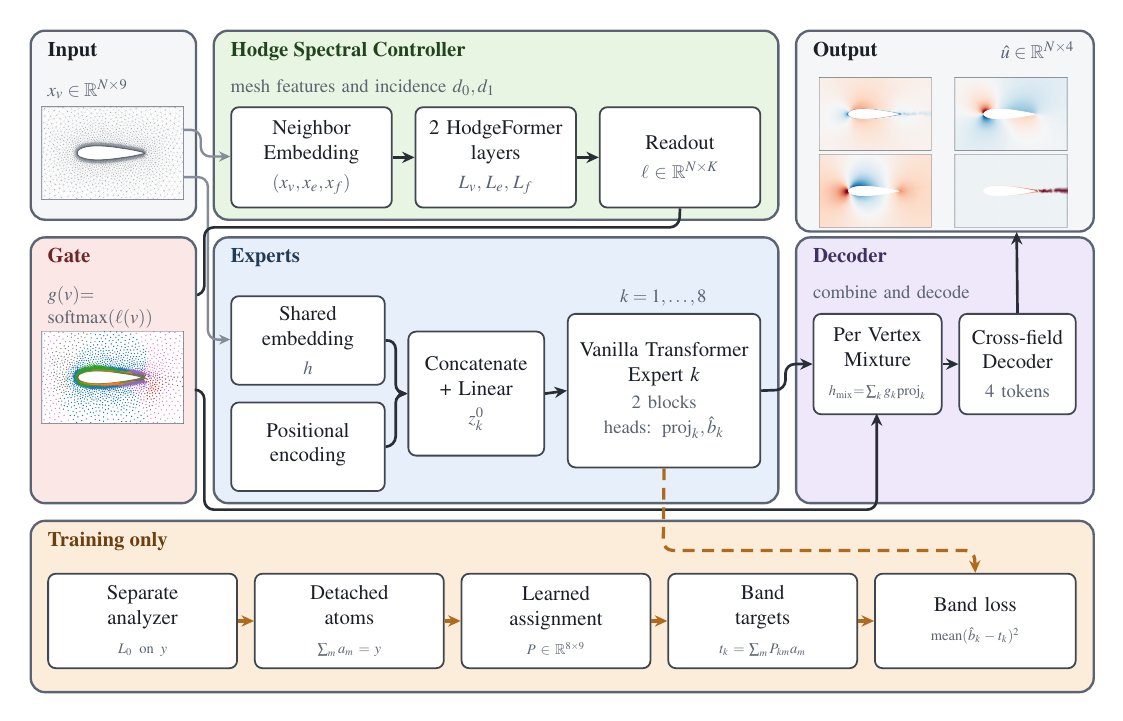}
  \fi
  \caption{Overview of the model. The Hodge controller routes transformer experts that combine shared embeddings with positional encodings at different frequency scales. Their representations are mixed and decoded into four fields. During training, a separate analyzer decomposes the reference fields into atoms, which a learned assignment combines into targets for the auxiliary expert heads (dashed path).}
  \label{fig:arch}
\ifdefined\overviewafterabstract
  \end{minipage}\par\medskip
\else
\end{figure}
\fi

\fi

\subsection{Simplicial complex and neighborhoods}
\label{sec:method-mesh}
The mesh is the simplicial complex $\mathcal M=(V,E,F)$, with $N=|V|$, $N_e=|E|$, and $N_f=|F|$ and signed incidence matrices $d_0\in\mathbb R^{N_e\times N}$ and $d_1\in\mathbb R^{N_f\times N_e}$, with non zero entries $-1$ or $+1$. The operators map vertex values to edges and edge values to faces, respectively. For each mesh element $i$, its neighborhood is the set of elements of the same type whose features contribute to the learned star at $i$. We call this set the \textit{operator support} because it specifies the entries that may be nonzero in row $i$ of the star matrix. The stored neighbor list is $\mathcal N^{\mathrm{full}}(i)$, while $\mathcal N(i)\subseteq\mathcal N^{\mathrm{full}}(i)$ denotes the subset used by the controller. Vertex and edge neighbor lists are initially generated from graph-hop rings, with at most eight hops. The controller uses a capped set of entries with $n_{vcap}=64$ for vertices $n_{ecap}=32$ edges. These lists are ordered by element index, so the retained entries are not necessarily the nearest neighbors. Truncation can also remove the reverse of a retained neighbor relation. Each face neighborhood contains only the face itself. Unused positions in the stored lists are padded with $-1$; these positions are not neighbors and do not contribute to the operator.

\subsection{Input features and input embeddings}
\label{sec:method-features}

The input feature are $x_v\in\mathbb R^{N\times c_v}$, $x_e\in\mathbb R^{N_e\times c_e}$, and $x_f\in\mathbb R^{N_f\times c_f}$. For vertices the input features can be denoted as follows:
\begin{equation}
x_v^{\mathrm{raw}}=[\mathrm{pos}^{\mathrm{raw}}_1,\mathrm{pos}^{\mathrm{raw}}_2,A_v^{\mathrm{dual}},u_{\mathrm{in}},v_{\mathrm{in}},\mathrm{SDF},n_x,n_y,\mathrm{is\_wall}],
  \label{eq:method-vertex-features}
\end{equation}
where $\mathrm{pos}^{\mathrm{raw}}_1,\mathrm{pos}^{\mathrm{raw}}_2$ are physical coordinate columns, $A_v^{\mathrm{dual}}$ is the barycentric dual area, computed as one third of the sum of incident triangle areas, and SDF denotes signed distance. The columns $u_{\mathrm{in}},v_{\mathrm{in}}$ contain inlet velocity components, $n_x,n_y$ contain normal components, and $\mathrm{is\_wall}$ is the wall indicator. 
Each vertex-input column, including the wall indicator, Z-score standardization takes place:
\begin{equation}
  (x_v)_{ij}=\frac{(x_v^{\mathrm{raw}})_{ij}-\mu_j^{\mathrm{in}}}{s_j^{\mathrm{in}}+10^{-8}},
  \qquad j=1,\ldots,c_v.
  \label{eq:method-input-standardization}
\end{equation}
,where $\mu_j^{\mathrm{in}}$ and $s_j^{\mathrm{in}}$ are computed by pooling vertices across the training meshes. The constant $10^{-8}$ prevents division by zero. The freestream speed is obtained from the inlet velocity components at the first vertex, and the dataset scales the velocity components $u_x,u_y$, the pressure $p$, and the turbulent kinematic viscosity $\nu_t$ by it:
\begin{equation}
U_\infty=\sqrt{u_{\mathrm{in}}^2+v_{\mathrm{in}}^2},
\qquad
\widetilde y=\left[\tfrac{u_x}{U_\infty},\tfrac{u_y}{U_\infty},\tfrac{p}{U_\infty^2},\tfrac{\nu_t}{U_\infty}\right],
\qquad
y_{vc}=\frac{\widetilde y_{vc}-\mu_c^{\mathrm{norm}}}{s_c^{\mathrm{norm}}+10^{-8}}.
  \label{eq:method-target-scaling}
\end{equation}
Here $v$ indexes vertices, $c$ indexes output channels, and $\mu_c^{\mathrm{norm}},s_c^{\mathrm{norm}}$ are the mean and standard deviation of the scaled training targets for channel c, computed across vertices and reused for all training and test cases. The standardized target is $y\in\mathbb R^{N\times4}$ and the model output is $\hat u\in\mathbb R^{N\times4}$. Physical reference fields are denoted by $y^{\mathrm{phys}}$. The turbulent kinematic viscosity target remains linear, without a logarithmic transformation. Dividing $\nu_t$ by $U_\infty$ leaves units of length. A dimensionless target therefore also requires a reference length, which is not explicit in this scaling.

The prediction branch maps the input features to input embeddings to one token per vertex through a shared two-layer perceptron with GELU activation. All eight experts receive the same token array $h$. 
\begin{equation}
  h=W_2\mathrm{GELU}(W_1x_v)\in\mathbb R^{N\times d},\qquad d=192.
  \label{eq:fwd-embed}
\end{equation}
The controller uses a separate neighbor embedding for vertex, edge, and face features and then maps each stream through its own three-layer perceptron, with ReLU after every linear layer. The resulting streams are $h_v\in\mathbb R^{N\times d_c}$, $h_e\in\mathbb R^{N_e\times d_c}$, and $h_f\in\mathbb R^{N_f\times d_c}$. Layer normalization is applied inside the subsequent PSD-Hodge layers.

\subsection{Hodge controller}
\label{sec:controller}

\textbf{Learned star operators.} The learned operators follow the approach in HodgeFormer \cite{nousias2026hodgeformer} that retains compositions of incidence and Hodge matrices while parameterizing the Hodge matrices through attention. 
Using the streams defined in \cref{sec:method-features}, each head first projects the normalized features to $\varphi_i=W_\varphi\mathrm{LN}(h_s)_i$, where $s\in\{v,e,f\}$ and $\varphi_i\in\mathbb R^{d_h}$. The controller has width $d_c=64$ and $H=4$ heads, so the per-head kernel-feature dimension is $d_h=d_c/H=16$.

These projected features determine the interaction between element $i$ and each neighbor $j\in\mathcal N(i)$ through a Gaussian kernel:
\begin{equation}
w_{ij}=\exp\left(-\frac{\|\varphi_i-\varphi_j\|_2^2}{2\sigma^2}\right),
\qquad \sigma=\operatorname{softplus}(\gamma)+10^{-8}.
  \label{eq:psd-kernel}
\end{equation}
Neighbors with similar projected features receive larger weights, while the learned bandwidth $\sigma$ controls how quickly the weights decrease with feature distance.

The matrix $\Omega$ collects these weights on the effective support and is zero elsewhere, including at padded entries. We normalize the weights using the row sums $D_i=\sum_j\Omega_{ij}$, with $D=\operatorname{diag}(D_i)$, and combine the resulting operator with the identity:
\begin{equation}
\hat S_{ij}=\frac{\Omega_{ij}}{\sqrt{D_iD_j+10^{-8}}},
\qquad \star x=(1-\beta)x+\beta\hat Sx,
\qquad \beta=\tfrac12\operatorname{sigmoid}(\alpha).
  \label{eq:psd-star}
\end{equation}
The identity term retains the element's own values, while $\hat Sx$ aggregates values from its neighbors. The learned coefficient $\beta$ controls their relative contribution and remains below $\tfrac12$.

The spectral properties of this construction depend on the neighborhood relation. A neighborhood is reciprocal when $j\in\mathcal N(i)$ implies $i\in\mathcal N(j)$. Under this condition, the ideal normalized matrix
$D^{-1/2}\Omega D^{-1/2}$ is symmetric and has eigenvalues in $[-1,1]$. The star formed with this ideal normalization is therefore symmetric positive-definite (SPD), with eigenvalues in $[1-2\beta,1]$.
This guarantee requires reciprocity of the effective support used by the operator. 
The configured caps and truncation can break reciprocity, for an asymmetric support, the identity contribution alone does not establish symmetry or the SPD property.

Face neighborhoods contain only the face itself. On a support of one element $w_{ii}=1$ and $\hat S=1$, so the Gaussian star of \cref{eq:psd-star} returns $x$ for every $\varphi$ and would not read the face features at all. The face star therefore uses a diagonal form that depends on them:
\begin{equation}
z_i=\frac{\|\varphi_i\|_2^2}{d_h\sigma^2},
\qquad (\star_2x)_i=(1-\beta+\beta e^{-z_i})x_i,
  \label{eq:method-face-star}
\end{equation}
This star scales each face independently, and its diagonal remains positive even if the exponential underflows
to zero. 
Stars labeled as inverse operators have their own learnable parameters and they are not calculated matrix inverses.

\textbf{Operator composition.} We denote the learned operators on vertices, edges, and faces by $L_v$, $L_e$, and $L_f$. For one attention head, let $x_v,x_e,x_f$ denote the normalized latent features supplied to that head.
\begin{equation}
\begin{aligned}
  L_v &:= \star_0^{-1}(x_v)\cdot d_0^T\cdot\star_1(x_e)\cdot d_0,\\
  L_e &:= d_0\cdot\star_0^{-1}(x_v)\cdot d_0^T\cdot\star_1(x_e)+\star_1^{-1}(x_e)\cdot d_1^T\cdot\star_2(x_f)\cdot d_1,\\
  L_f &:= d_1\cdot\star_1^{-1}(x_e)\cdot d_1^T\cdot\star_2(x_f).
\end{aligned}
\label{eq:ctrl-layer}
\end{equation}
\label{eq:method-edge-face-operators}
The vertex operator reads vertex and edge features, the edge operator reads all three feature arrays, and the face operator reads edge and face features.  $\star_0^{-1}$ and $\star_1^{-1}$ have separate learnable parameters. The two terms of $L_e$ act on the same projected edge values and their outputs are added. The head outputs are concatenated and projected back to the embedding dimension, as described below.

If the stars in $L_v$ are symmetric positive definite, then $Q=d_0^T\star_1(x_e)d_0$ is symmetric positive semidefinite and
\begin{equation}
  L_v=\star_0^{-1}(x_v)Q
  \sim\star_0^{-1/2}(x_v)Q\star_0^{-1/2}(x_v).
\label{eq:l0-spectrum}
\end{equation}
Here $\star_0^{-1/2}(x_v)$ denotes the symmetric square root of the independently learned $\star_0^{-1}(x_v)$. Under this assumption, $L_v$ has a real, nonnegative spectrum, although it is generally not symmetric in the Euclidean inner product. 

\textbf{Overall Controller Architecture.} The controller applies two PSD-Hodge layers followed by a vertex readout. Each layer has four heads of width $16$ and feedforward width $256$. After the second PSD-Hodge layer, the controller reads only $h_v$ and computes
\begin{equation}
  \ell=\mathrm{Linear}_{64\to8}
    (\mathrm{GELU}(\mathrm{Linear}_{64\to64}(\mathrm{LN}(h_v))))
    \in\mathbb R^{N\times K},\qquad K=8.
\label{eq:method-controller-readout}
\end{equation}
The mixture module converts these logits into routing weights through a softmax without temperature scaling. With the batch axis included, the logit tensor has shape $(B_{\mathrm{batch}},N,K)$.

\subsection{Transformer experts and decoder}
\label{sec:experts}

Expert $k$ receives the token array $h$ joined with Fourier features of $\mathrm{pos}\in\mathbb R^{N\times2}$, the two standardized coordinate columns of $x_v$:
\begin{equation}
  z^0_k=W^k_{\text{in}}\big[\,h\;;\;\sin(2\pi\,\mathrm{pos}\,B_k)\;;\;\cos(2\pi\,\mathrm{pos}\,B_k)\,\big]\in\mathbb R^{N\times d_x},
  \qquad B_k=\sigma_kG_k\in\mathbb R^{2\times n_F},
  \label{eq:method-expert-input}
\end{equation}
where the semicolon joins arrays along the feature dimension, the expert width is $d_x=88$, the number of frequencies is $n_F=8$, and the entries of $G_k$ are drawn once from a standard normal distribution and then held fixed. All projections in this subsection include bias terms, which are omitted from the equations. The scales are geometric, $(\sigma_1,\ldots,\sigma_K)=(1.0,1.7,2.8,4.8,8.0,13.5,22.6,38.0)$, and are the only difference between the experts at initialization.

Each expert applies $L_x=2$ transformer blocks to $z^0_k$. A block applies multi-head softmax self-attention over all $N$ vertices, then a feedforward network to each vertex separately, with layer normalization before each sublayer and a residual connection around it. The attention uses $h_x=8$ heads of width $d_x/h_x=11$ and the feedforward width is $4d_x=352$. The experts receive no incidence matrices and no neighborhood indices. Two linear maps read the output $z_k$ of the last block: $\operatorname{proj}_k$, which returns an array of width $d$, and $W^k_b$, which returns the auxiliary prediction $\hat b_k\in\mathbb R^{N\times4}$. The auxiliary prediction is supervised against the per-expert target constructed in \cref{sec:analyzer} and is not used to form the field prediction.

All $K$ experts are evaluated at every vertex. The controller logits of \cref{eq:method-controller-readout} pass through a softmax, which gives weights $\gate(v)\in\simplex$ that are nonnegative and sum to one, and the expert representations are combined as
\begin{equation}
  h_{\text{mix}}(v)=\sum_{k=1}^{K}\gate_k(v)\operatorname{proj}_k(z_k)(v)\in\mathbb R^{d}.
  \label{eq:method-mixture}
\end{equation}
The weights set the combination and not the number of experts evaluated, so routing does not reduce expert computation.

The decoder applies layer normalization and dropout to $h_{\text{mix}}$, giving $m(v)\in\mathbb R^{d}$, and forms one token per output channel by adding a row $E_{\mathrm{field},c}$ of a learned field embedding $E_{\mathrm{field}}\in\mathbb R^{4\times d}$:
\begin{equation}
  t^0_{vc}=m(v)+E_{\mathrm{field},c},\qquad c=1,\ldots,4.
  \label{eq:method-decoder-token-init}
\end{equation}
Two blocks of the same form act on the four tokens of one vertex, with width $d=192$, four heads of width $48$, and feedforward width $768$. The attention axis has length four, so the attention relates the output fields at a fixed vertex and does not mix vertices. Writing $t^2_{vc}$ for the token after the second block, a readout shared by all vertices and fields returns one scalar per field,
\begin{equation}
  \hat u_{vc}=\operatorname{Linear}_{64\to1}\!\left(\mathrm{GELU}\!\left(\operatorname{Linear}_{192\to64}(\mathrm{LN}(t^2_{vc}))\right)\right),
  \label{eq:method-decoder-output}
\end{equation}
which gives $\hat u\in\mathbb R^{N\times4}$. \Cref{sec:appendix-expert-fourier-bandwidths} reports what the Fourier scales do and do not constrain.

\subsection{Auxiliary targets}
\label{sec:analyzer}

Each expert produces an auxiliary prediction $\hat b_k$ alongside its contribution to the field. The targets for those predictions are built from the reference fields themselves. The reference field is separated into components that vary over different distances on the mesh, and each expert is assigned a combination of them. The separation is produced by repeated local averaging. We define averaging weights on every edge as conductances. The final prediction is trained against the original reference fields and not against these components.

\textbf{Conductances.} A conductance $w_e>0$ is the weight that edge $e$ carries when a vertex value is replaced by an average over its neighbors. A weight is needed because the mesh cells differ in size, so the number of edges between two vertices is not proportional to the distance between them, and averaging over the unweighted graph would cover different amounts of physical space in different regions of the same mesh. Weighting each edge separates the two. The name follows the electrical reading of the quadratic form given below, in which $w_e$ is the conductance of a resistor joining the endpoints of $e$.

The conductances are not fixed in advance, but an edge network $\mathrm{MLP}_e$ maps the $c_e$ features of each edge to a scalar through a hidden layer of width $64$ with GELU activation, and softplus with an added floor $\epsilon_w$ makes the result positive:
\begin{equation}
w_e=\operatorname{softplus}(\mathrm{MLP}_e(x_e))+\epsilon_w,\qquad \epsilon_w=10^{-4}.
  \label{eq:method-analyzer-conductances}
\end{equation}
The conductances then define two diagonal operators and a Laplacian:
\begin{equation}
\star_1=\operatorname{diag}(w),\qquad \mu=|d_0|^\top w,\qquad
\star_0=\operatorname{diag}(\mu),\qquad L_0=\star_0^{-1}d_0^\top\star_1d_0.
  \label{eq:an-star}
\end{equation}
Here $w$ collects the edge conductances and $\mu_v=\sum_{e\ni v}w_e$ is the total conductance of the edges meeting vertex $v$. The vertex operator is therefore determined by the conductances, and no second network is used for it. Both operators are diagonal, and they are separate from the controller stars of \cref{sec:controller}. We use signed incidence matrices with nonzero entries $-1$ and $+1$. For any positive conductances
\[
  x^\top d_0^\top\star_1d_0x=\sum_{e=(i,j)}w_e(x_i-x_j)^2\ge0,
\]
so $L_0$ is positive semidefinite by construction. The implementation bounds vertex masses below by $\epsilon_w$, which changes nothing at a vertex with incident edges because every conductance is already at least $\epsilon_w$. Taking the vertex operator to be the total incident conductance, rather than a second learned quantity, is what bounds the spectrum. With this choice $L_0$ is the random-walk Laplacian, that is, the operator returning the difference between a vertex value and the conductance-weighted average of its neighbors, and its eigenvalues lie in $[0,2]$ for any positive conductances. A separately learned vertex mass removes that bound, and on these graded meshes the largest eigenvalue then grows by several orders of magnitude, so no affordable number of smoothing steps reaches the slowly varying part of the field.

\textbf{The solution at different spatial scales.} One smoothing step replaces each vertex value by a mixture of itself and the conductance-weighted average of its neighbors,
\[
  \bar y_i\leftarrow\tfrac12\bar y_i+
  \tfrac12\sum_{j\sim i}\frac{w_{ij}}{\mu_i}\bar y_j,
\]
which is $\bar y\leftarrow(I-\tau L_0)\bar y$ with $\tau=\tfrac12$. The eigenvalue bound above makes this step stable for every sample without estimating the largest eigenvalue, and a fixed step keeps the construction reproducible from run to run. Repeating the step $m$ times mixes each vertex with the vertices reachable within about $m$ edges, so the number of steps selects a distance on the mesh.
How much of each component of the field survives $m$ steps is set by the eigenvalues of $L_0$. $L_0$ is self-adjoint and positive semidefinite under the inner product $\langle f,g\rangle_\mu=\sum_v\mu_vf_vg_v$. It is generally not symmetric in Euclidean coordinates, but $\star_0^{1/2}L_0\star_0^{-1/2}$ is a symmetric positive-semidefinite normalized Laplacian, so the eigenvalues of $L_0$ are real and lie in $[0,2]$. A component with eigenvalue $\lambda$ is multiplied by $1-\lambda/2$ at each step and by $(1-\lambda/2)^m$ after $m$ steps. A constant field has $\lambda=0$ and is unchanged. Components that alternate across strongly conducting edges have the largest eigenvalues and are removed first. What remains after $m$ steps is the part of the field that varies slowly relative to the conductances, and what was removed is recovered as a difference between two smoothing levels. We retain the smoothed fields at $L=8$ step counts,
\begin{equation}
(m_0,\ldots,m_{L-1})=(2,4,6,8,12,16,24,32),
\qquad \bar y^{(j)}=(I-\tau L_0)^{m_j}y,\quad j=0,\ldots,L-1,
  \label{eq:method-analyzer-ladder}
\end{equation}
and form $M=L+1=9$ components from the most smoothed field, the differences between consecutive retained levels, and the difference between the reference field and the least smoothed field. We call these components atoms:
\begin{equation}
a_0=\bar y^{(L-1)},\qquad
a_i=\bar y^{(L-1-i)}-\bar y^{(L-i)}\quad(1\le i\le L-1),
\qquad a_{M-1}=y-\bar y^{(0)}.
  \label{eq:an-atoms}
\end{equation}
Here $a_0$ is the field after $32$ steps and holds what survives the most averaging. Each intermediate atom holds the variation removed between two consecutive retained levels, and $a_{M-1}$ holds the variation removed by the first two steps. The differences cancel when the atoms are added, so $\sum_m a_m=y$ up to floating-point rounding, whatever the conductances are. The atoms are not an orthogonal decomposition. Their spectral responses overlap, so they do not divide the field into disjoint frequency bands.

\textbf{Assignment.} There are $M=9$ atoms and $K=8$ experts, and the target of each expert is a weighted sum of the atoms. We learn the weights. For each expert--atom pair we hold a score $\theta\in\mathbb R^{K\times M}$, and a softmax across experts converts the scores of each atom into nonnegative weights:

\begin{equation}
P_{km}=\frac{\exp(\theta_{km}/T)}{\sum_{k'=1}^{K}\exp(\theta_{k'm}/T)},
\qquad T=1,\qquad
t_k=\sum_{m=0}^{M-1}P_{km}a_m.
  \label{eq:an-assign}
\end{equation}

The weight $P_{km}$ specifies the fraction of atom $a_m$ included in expert $k$'s target $t_k$. The same assignment matrix is used at every vertex and for every case. It is separate from the controller's input-dependent routing weights. Each column of $P$ sums to one, so every atom is fully distributed among the expert targets and $\sum_{k=1}^{K}t_k=y$ algebraically. Each expert's auxiliary output $\hat b_k$ predicts the four fields in its assigned target $t_k$.

During training, the atoms are treated as constants when forming the targets, an operation called detachment. The training loss can therefore update the assignment scores without passing gradients through the atoms to the analyzer conductances. The assignment can combine several atoms in one expert target or distribute one atom across several targets. No constraint orders the expert targets from low to high frequency.

The assignment constructor sets a larger score for each pair $(k,a_{k-1})$, with $k=1,\ldots,K$, and sets equal scores across experts for $a_{M-1}$. The subsequent global Xavier initialization replaces these values, so training begins with Xavier-initialized assignment scores.

\subsection{Training Objective}
\label{sec:method-objective}

We train the final prediction to match the reference fields and each expert's auxiliary output to match its assigned target. Two additional penalties encourage balanced average routing weights and discourage assignments that give an expert a target with a very small share of the total squared target norm. The combined objective is
\begin{equation}
\mathcal L=\mathcal L_{\mathrm{vol}}+\mathcal L_{\mathrm{surf}}+\lambda_b\operatorname{mean}_{v,c,k}(\hat b_{vck}-t_{vck})^2+\lambda_{\mathrm{bal}}K\sum_{k=1}^K\big(\bar g_k-\tfrac1K\big)^2+\lambda_{\mathrm{floor}}\sum_{k=1}^K[\mathrm{ReLU}(\rho-p_k)]^2.
\label{eq:method-loss}
\end{equation}
Here $\mathcal L_{\mathrm{vol}}$ and $\mathcal L_{\mathrm{surf}}$ are mean squared errors of the normalized field residual over the non-wall and the wall vertices, $\bar g_k$ is expert $k$'s routing weight averaged over the batch and vertices, and $p_k$ is its assigned target's share of the total squared target norm. The three are defined in \cref{sec:appendix-objective}. We use $(\lambda_b,\lambda_{\mathrm{bal}},\lambda_{\mathrm{floor}})=(0.5,0.01,1.0)$ and $\rho=0.02$.

\newcommand{\arxivqualinresults}{}
\section{Experiments}
\label{sec:exp}

\subsection{Simulation Setup}
\label{sec:setup}

AirFRANS~\citep{bonnet2022airfrans} provides vertex positions, input features, and four solution fields. We generate anisotropic triangular working meshes with gmsh~\citep{geuzaine2009gmsh}, each with approximately $4.9$k vertices and refinement near the airfoil, and interpolate the reference fields onto them. The data is two-dimensional, and coordinates are embedded at $z=0$. The nine vertex inputs are defined in \cref{sec:method-features}. They encode geometry, inlet velocity, signed distance, the normals supplied by the dataset, and wall membership. Boundary conditions enter through these features. The wall indicator selects the vertices that form the surface term of the training objective and the wall error metric defined below. Edge features determine the analyzer conductances. The reported controller updates vertices alone and reads edge features in its vertex operator; the VEF variant updates edges and faces as well.

We use the official $800/200$ split. Each case has up to eight remeshes, and one of them is drawn at random every time the case is read during training; evaluation uses one fixed mesh per case. Targets are scaled by the sample's freestream speed and standardized using training statistics(\cref{eq:method-target-scaling}).

Per-field relative $L_2$ and $R^2$ are evaluated in physical units. Volume and Surf are mean squared errors of the physical prediction error divided by the per-field standard deviation of the physical targets over the training set, taken over the non-wall and the wall vertices respectively. We also report the relative error of the lift coefficient and its Spearman correlation across test cases. All models are evaluated against interpolated targets on the same working meshes, so these numbers are not directly comparable with benchmark scores computed on the original meshes.

Our checkpoint is selected from $20$ checkpoints by test-set error, whereas the baselines use validation-based selection. Our models train on all $800$ training cases, while the baselines reserve $80$ for validation. These protocol differences limit attribution of the reported differences to architecture alone. Training uses AdamW~\citep{loshchilov2019adamw} with a one-cycle schedule~\citep{smith2019onecycle} for $400$ epochs and one mesh per batch. Autocast runs in \code{bfloat16}, while the incidence and analyzer operations stay in \code{fp32}. Each run uses one NVIDIA RTX~4090 with $24$\,GB of memory. Full settings are given in \cref{sec:appendix-config}.

\begin{table}[!tb]
\caption{\textbf{Comparison on the official AirFRANS $800/200$ split.} Every row is evaluated on the
same $\approx4.9$k-vertex working meshes and evaluation metrics. Baselines use their full input-feature arms; training and checkpoint-selection differences are specified in \cref{sec:setup}. \emph{Mesh use} names how mesh structure
reaches the prediction: \emph{none} for a point-set model, \emph{operator} when a learned Hodge
Laplacian computes the fields, \emph{router} when it computes only the mixing weights. $\CL$ is the
relative error of the lift over the $200$ test cases and $\rhoL$ its Spearman correlation
with the reference.}
\label{tab:main}
\centering\footnotesize
\setlength{\tabcolsep}{5pt}
\begin{tabular*}{\textwidth}{@{\extracolsep{\fill}}lrccccccc@{}}
\toprule
model & params & mesh use & $\relL\downarrow$ & $R^2\uparrow$ & Volume$\downarrow$ & Surf$\downarrow$ & $\CL\downarrow$ & $\rhoL\uparrow$ \\
\midrule
Transolver~\citep{wu2024transolver}          & $6.00$M & none     & 0.0436 & 0.9953 & 0.00265 & \best{0.00143} & \best{0.0103} & \best{0.99989} \\
PointNet~\citep{qi2017pointnet}              & $0.08$M & none     & 0.0689 & 0.9904 & 0.00451 & 0.00440 & 0.0149 & 0.99965 \\
Pointwise MLP                                & $0.02$M & none     & 0.1743 & 0.9404 & 0.03576 & 0.03969 & 0.0860 & 0.99338 \\
PSD-star HodgeFormer~\citep{nousias2026hodgeformer} & $0.47$M & operator & 0.0800 & 0.9885 & 0.01028 & 0.01048 & 0.0681 & 0.99883 \\
\textbf{Hodge-routed MoE (ours)}             & $3.02$M & router   & \best{0.0300} & \best{0.9983} & \best{0.00140} & 0.00164 & 0.0225 & 0.99988 \\
\bottomrule
\end{tabular*}
\end{table}

\subsection{Results}
\label{sec:results}

\Cref{tab:main} compares the models on the official split. Our model has the lowest relative $L_2$
error and the lowest interior mean squared error of the five, $31\%$ and $47\%$ below Transolver at
half its parameter count. Transolver has the lower error on the two wall quantities, by a factor of
$1.14$ on the surface error and $2.2$ on the lift coefficient. The fourth row is a HodgeFormer
surrogate, which keeps the learned Hodge Laplacian in the prediction path. The per-field breakdown
is given in full in the supplementary material: our model has the lowest relative $L_2$ error on
three of the four fields and the second lowest on $\uy$, and the largest difference is on the eddy
viscosity, the least accurate output of every model, where our error is $46\%$ below Transolver's.
The aggregate is the mean over the four fields. The fixed reference decomposition is illustrated for pressure in \cref{fig:bands}. The design was selected through ablations over the source of the
gate, the width of the predictor, fixed against learned band boundaries, the guard against an empty
band, and the number of cochain degrees the controller operates on, given in the supplementary material.


\Cref{fig:qual} shows one test case at the reported checkpoint, with the reference, prediction, and absolute error for each field. Its per-field relative $L_2$ errors are $0.0047$, $0.0260$, $0.0281$, and $0.0676$, compared with test-set means of $0.0056$, $0.0259$, $0.0255$, and $0.0629$. Visible errors occur near the airfoil and along the turbulent-viscosity wake.
\begin{figure}[!ht]
  \centering
  \begin{tikzpicture}
    \node[anchor=south west,inner sep=0pt] (qual)
      {\includegraphics[width=0.92\textwidth]{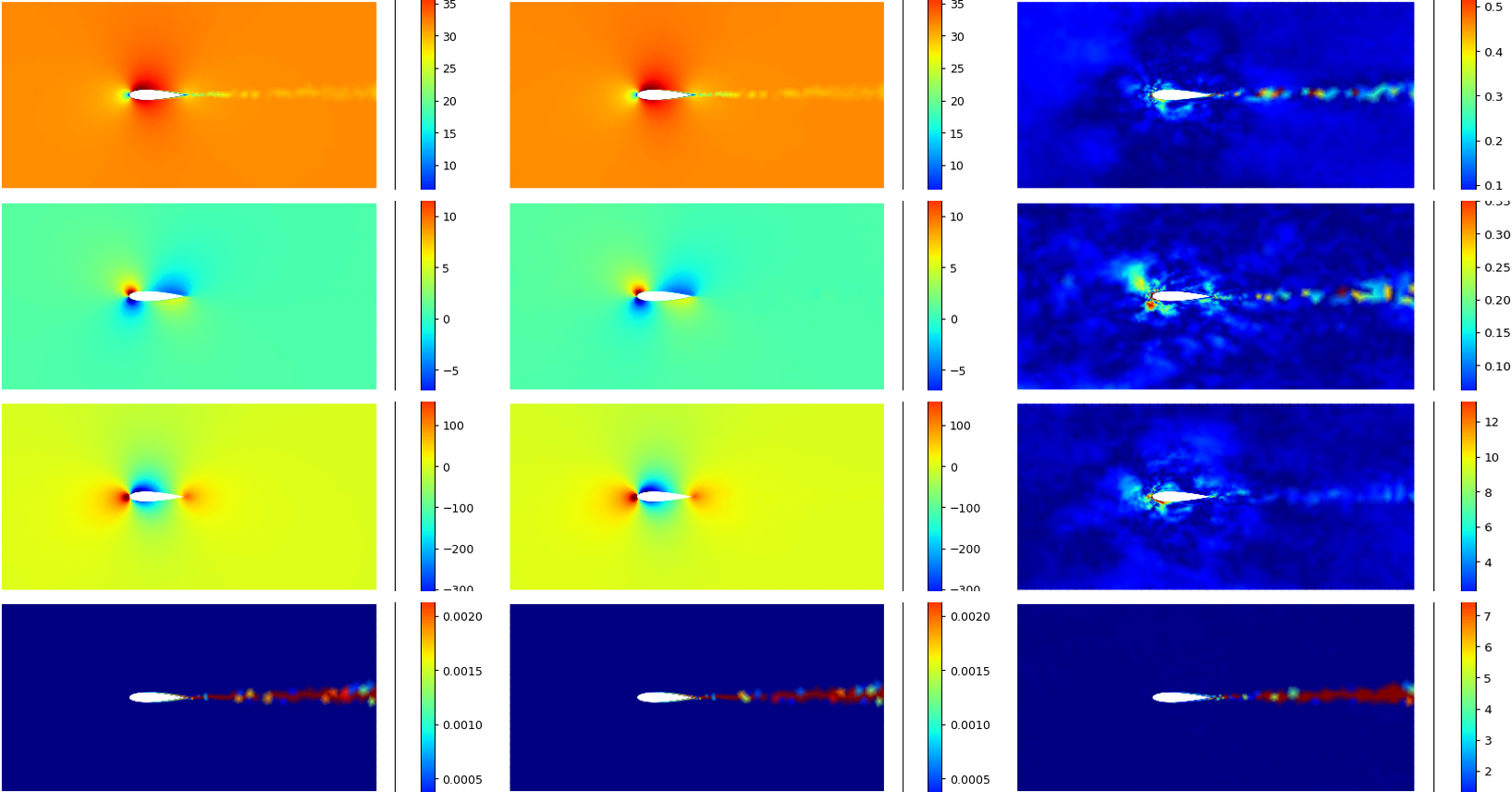}};
    \begin{scope}[x={(qual.south east)},y={(qual.north west)}]
      \node[anchor=south,font=\small] at (0.1248,1) {Reference};
      \node[anchor=south,font=\small] at (0.4606,1) {Prediction};
      \node[anchor=south,font=\small] at (0.8040,1) {Absolute error};
      \node[anchor=south,font=\small,rotate=90] at (0,0.8803) {$\ux$};
      \node[anchor=south,font=\small,rotate=90] at (0,0.6262) {$\uy$};
      \node[anchor=south,font=\small,rotate=90] at (0,0.3732) {$p$};
      \node[anchor=south,font=\small,rotate=90] at (0,0.1197) {$\nut$};
    \end{scope}
  \end{tikzpicture}
  \caption{\textbf{One test case at the reported checkpoint (epoch $380$).} Columns are the reference,
  the prediction and the absolute error; rows are $\ux$, $\uy$, $p$ and $\nut$. All panels are on the
  model's own working mesh in physical units, with the reference and the prediction sharing each
  field's colour scale. The per-field relative $L_2$ for this case is $0.0047$, $0.0260$, $0.0281$ and
  $0.0676$. Test-set means are listed in \cref{tab:appendix-perfield}; the panel illustrates one case rather than a test-set average.}
  \label{fig:qual}
\end{figure}

\begin{figure}[!ht]
  \centering
  \includegraphics[width=\textwidth]{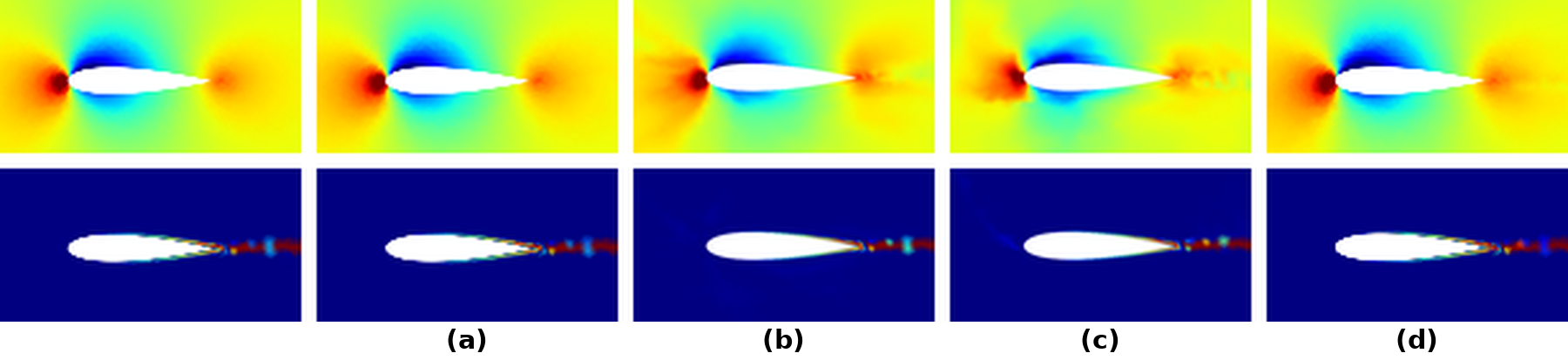}
  \par\vspace{2pt}
  \noindent\makebox[0.2\textwidth][c]{\shortstack{\scriptsize CFD reference\\\scriptsize ---}}%
  \makebox[0.2\textwidth][c]{\shortstack{\scriptsize $\mathrm{rel}\text{-}L_2(p)=0.026$\\\scriptsize $\mathrm{rel}\text{-}L_2(\nut)=0.063$}}%
  \makebox[0.2\textwidth][c]{\shortstack{\scriptsize $\mathrm{rel}\text{-}L_2(p)=0.030$\\\scriptsize $\mathrm{rel}\text{-}L_2(\nut)=0.116$}}%
  \makebox[0.2\textwidth][c]{\shortstack{\scriptsize $\mathrm{rel}\text{-}L_2(p)=0.066$\\\scriptsize $\mathrm{rel}\text{-}L_2(\nut)=0.144$}}%
  \makebox[0.2\textwidth][c]{\shortstack{\scriptsize $\mathrm{rel}\text{-}L_2(p)=0.085$\\\scriptsize $\mathrm{rel}\text{-}L_2(\nut)=0.155$}}
  \caption{Pressure $p$ (top) and turbulent viscosity $\nut$ (bottom) on the reference colour scales; left to right, the CFD reference, (a) our model, (b) Transolver, (c) PointNet, and (d) the best PSD-star HodgeFormer. Values beneath each model panel are mean relative $L_2$ errors over the $200$ test cases. The models differ more on $\nut$ than on $p$; the reference panel shows fine-scale structure in the $\nut$ wake.}
  \label{fig:compare}
  \vspace{-1em}
\end{figure}

\Cref{fig:compare} places the four models beside the CFD reference on a common colour scale. The
models separate more on the turbulent viscosity than on the pressure: the mean relative $L_2$ over
the $200$ test cases is $0.063$, $0.116$, $0.144$ and $0.155$ on $\nut$ for (a)--(d), against
$0.026$, $0.030$, $0.066$ and $0.085$ on $p$. The reference carries fine-scale structure in the
$\nut$ wake, and the ordering of the panels on that field follows the ordering of those errors.

\section{Discussion and Conclusion}
\label{sec:discussion}

We propose a Hodge-routed mixture of transformer experts for steady-PDE prediction on unstructured meshes. A learned Hodge Laplacian computes the routing weights, and a separate analyzer constructs multiscale supervision targets. On the AirFRANS working mesh, mean relative $L_2$ and interior mean squared errors decrease by $31\%$ and $47\%$ relative to Transolver, from $0.0436$ to $0.0300$ and from $0.00265$ to $0.00140$, respectively. The largest per-field improvement is in turbulent viscosity, whose relative $L_2$ error decreases by approximately $46\%$, from $0.1158$ to $0.0629$ (\cref{tab:appendix-perfield}). Introducing the Hodge controller into the learned-band mixture reduces mean relative $L_2$ error from $0.0345$ to $0.0300$ (\cref{tab:appendix-main}), with larger additional gains in velocity and pressure than in turbulent viscosity (\cref{tab:appendix-perfield}).

The mean relative $L_2$ error for pressure is lower than Transolver's, while surface and relative lift-coefficient errors are higher: $0.00164$ versus $0.00143$ and $0.0225$ versus $0.0103$, respectively (\cref{tab:main}). Both models retain lift rank correlations above $0.9998$, preserving case ordering more accurately than force magnitudes. Pressure errors along the airfoil contour therefore require further examination.

We follow Transolver's evaluation format on the official AirFRANS $800/200$ split, with one training run per configuration and evaluation on a remeshed working mesh of approximately $4.9$K vertices. The results do not establish accuracy at the original CFD resolution or robustness across random seeds. The analyzer's algebraic target reconstruction holds independently of the controller's conditional spectral guarantee. The implementation does not enforce the reciprocal neighborhoods required by that guarantee, and prediction accuracy does not validate it. Every expert is evaluated, and dense attention scales quadratically with mesh size.

\bibliographystyle{unsrtnat}
\bibliography{refs}

@String(CVPR  = {IEEE Conf. Comput. Vis. Pattern Recog.})

@String(NeurIPS = {Adv. Neural Inform. Process. Syst.})

@String(ICML  = {Int. Conf. Mach. Learn.})

@String(ICLR  = {Int. Conf. Learn. Represent.})

@String(AAAI  = {AAAI})

@String(JMLR  = {J. Mach. Learn. Res.})

@String(TMLR  = {Trans. Mach. Learn. Res.})

@String(TOG   = {ACM Trans. Graph.})

@inproceedings{nousias2026hodgeformer,
  author    = {Nousias, Akis and Nousias, Stavros},
  title     = {{HodgeFormer}: Transformers for Learnable Operators on Triangular Meshes through Data-Driven Hodge Matrices},
  booktitle = {IEEE/CVF Winter Conf. on Applications of Computer Vision (WACV)},
  year      = {2026},
  note      = {arXiv:2509.01839}
}

@inproceedings{bonnet2022airfrans,
  author    = {Bonnet, Florent and others},
  title     = {{AirFRANS}: High Fidelity Computational Fluid Dynamics Dataset for Approximating {Reynolds}-Averaged {Navier--Stokes} Solutions},
  booktitle = NeurIPS,
  year      = {2022}
}

@inproceedings{wu2024transolver,
  author    = {Wu, Haixu and Luo, Huakun and Wang, Haowen and Wang, Jianmin and Long, Mingsheng},
  title     = {Transolver: A Fast Transformer Solver for {PDEs} on General Geometries},
  booktitle = ICML,
  year      = {2024}
}

@inproceedings{li2021fourier,
  author    = {Li, Zongyi and others},
  title     = {Fourier Neural Operator for Parametric Partial Differential Equations},
  booktitle = ICLR,
  year      = {2021}
}

@article{li2023fourier,
  author    = {Li, Zongyi and Huang, Daniel Zhengyu and Liu, Burigede and Anandkumar, Anima},
  title     = {Fourier Neural Operator with Learned Deformations for {PDEs} on General Geometries},
  journal   = JMLR,
  volume    = {24},
  number    = {388},
  pages     = {1--26},
  year      = {2023}
}

@inproceedings{li2023geometry,
  author    = {Li, Zongyi and Kovachki, Nikola and Choy, Chris and Li, Boyi and Kossaifi, Jean and Otta, Shourya and Nabian, Mohammad Amin and Stadler, Maximilian and Hundt, Christian and Azizzadenesheli, Kamyar and Anandkumar, Anima},
  title     = {Geometry-Informed Neural Operator for Large-Scale {3D} {PDEs}},
  booktitle = NeurIPS,
  year      = {2023}
}

@inproceedings{hao2023gnot,
  author    = {Hao, Zhongkai and others},
  title     = {{GNOT}: A General Neural Operator Transformer for Operator Learning},
  booktitle = ICML,
  year      = {2023}
}

@inproceedings{cao2021choose,
  author    = {Cao, Shuhao},
  title     = {Choose a Transformer: Fourier or {Galerkin}},
  booktitle = NeurIPS,
  year      = {2021}
}

@inproceedings{pfaff2021learning,
  author    = {Pfaff, Tobias and others.},
  title     = {Learning Mesh-Based Simulation with Graph Networks},
  booktitle = ICLR,
  year      = {2021}
}

@inproceedings{li2020neural,
  author    = {Li, Zongyi and others},
  title     = {Neural Operator: Graph Kernel Network for Partial Differential Equations},
  booktitle = {ICLR Workshop on Integration of Deep Neural Models and Differential Equations},
  year      = {2020}
}

@article{kovachki2023neural,
  author    = {Kovachki, Nikola and Li, Zongyi and Liu, Burigede and Azizzadenesheli, Kamyar and Bhattacharya, Kaushik and Stuart, Andrew and Anandkumar, Anima},
  title     = {Neural Operator: Learning Maps Between Function Spaces with Applications to {PDEs}},
  journal   = JMLR,
  volume    = {24},
  number    = {89},
  pages     = {1--97},
  year      = {2023},
  doi       = {10.5555/3648699.3648788}
}

@article{li2024physics,
  author    = {Li, Zongyi and Zheng, Hongkai and Kovachki, Nikola and Jin, David and Chen, Haoxuan and Liu, Burigede and Azizzadenesheli, Kamyar and Anandkumar, Anima},
  title     = {Physics-Informed Neural Operator for Learning Partial Differential Equations},
  journal   = {ACM/IMS J. Data Sci.},
  year      = {2024},
  doi       = {10.1145/3648506}
}

@article{azizzadenesheli2023neural,
  author    = {Azizzadenesheli, Kamyar and Kovachki, Nikola and Li, Zongyi and Liu-Schiaffini, Miguel and Kossaifi, Jean and Anandkumar, Anima},
  title     = {Neural Operators for Accelerating Scientific Simulations and Design},
  journal   = {arXiv preprint arXiv:2309.15325},
  year      = {2023}
}

@article{lu2021learning,
  author  = {Lu, Lu and Jin, Pengzhan and Pang, Guofei and Zhang, Zhongqiang and Karniadakis, George Em},
  title   = {Learning Nonlinear Operators via {DeepONet} Based on the Universal Approximation Theorem of Operators},
  journal = {Nature Machine Intelligence},
  volume  = {3},
  number  = {3},
  pages   = {218--229},
  year    = {2021},
  doi     = {10.1038/s42256-021-00302-5}
}

@article{raissi2019physics,
  author  = {Raissi, Maziar and Perdikaris, Paris and Karniadakis, George Em},
  title   = {Physics-Informed Neural Networks: A Deep Learning Framework for Solving Forward and Inverse Problems Involving Nonlinear Partial Differential Equations},
  journal = {J. Comput. Phys.},
  volume  = {378},
  pages   = {686--707},
  year    = {2019},
  doi     = {10.1016/j.jcp.2018.10.045}
}

@article{karniadakis2021physics,
  author  = {Karniadakis, George Em and Kevrekidis, Ioannis G. and Lu, Lu and Perdikaris, Paris and Wang, Sifan and Yang, Liu},
  title   = {Physics-Informed Machine Learning},
  journal = {Nature Reviews Physics},
  volume  = {3},
  number  = {6},
  pages   = {422--440},
  year    = {2021}
}

@article{vinuesa2022enhancing,
  author  = {Vinuesa, Ricardo and Brunton, Steven L.},
  title   = {Enhancing Computational Fluid Dynamics with Machine Learning},
  journal = {Nature Computational Science},
  volume  = {2},
  number  = {6},
  pages   = {358--366},
  year    = {2022}
}

@article{brunton2020machine,
  author  = {Brunton, Steven L. and Noack, Bernd R. and Koumoutsakos, Petros},
  title   = {Machine Learning for Fluid Mechanics},
  journal = {Annual Review of Fluid Mechanics},
  volume  = {52},
  pages   = {477--508},
  year    = {2020}
}

@inproceedings{qi2017pointnet,
  author    = {Qi, Charles R. and Su, Hao and Mo, Kaichun and Guibas, Leonidas J.},
  title     = {{PointNet}: Deep Learning on Point Sets for {3D} Classification and Segmentation},
  booktitle = CVPR,
  year      = {2017}
}

@article{li2022transformer,
  author  = {Li, Zijie and Meidani, Kazem and Farimani, Amir Barati},
  title   = {Transformer for Partial Differential Equations' Operator Learning},
  journal = TMLR,
  year    = {2022},
  note    = {arXiv:2205.13671}
}

@article{li2023scalable,
  author  = {Li, Zijie and Shu, Dule and Farimani, Amir Barati},
  title   = {Scalable Transformer for {PDE} Surrogate Modeling},
  journal = {arXiv preprint arXiv:2305.17560},
  year    = {2023}
}

@article{wen2025geometry,
  author  = {Wen, Shizheng and Kumbhat, Arsh and Lingsch, Levi and Mousavi, Sepehr and Zhao, Yizhou and Chandrashekar, Praveen and Mishra, Siddhartha},
  title   = {Geometry Aware Operator Transformer as an Efficient and Accurate Neural Surrogate for {PDEs} on Arbitrary Domains},
  journal = {arXiv preprint arXiv:2505.18781},
  year    = {2025}
}

@article{bleeker2025abupt,
  author  = {Bleeker, Maurits and Dorfer, Matthias and Kronlachner, Tobias and Sonnleitner, Reinhard and Alkin, Benedikt and Brandstetter, Johannes},
  title   = {{AB-UPT}: Scaling Neural {CFD} Surrogates for High-Fidelity Automotive Aerodynamics Simulations via Anchored-Branched Universal Physics Transformers},
  journal = TMLR,
  year    = {2025}
}

@inproceedings{lee2024inducing,
  author    = {Lee, Seungjun and Oh, Taegon},
  title     = {Inducing Point Operator Transformer: A Flexible and Scalable Architecture for Solving {PDEs}},
  booktitle = AAAI,
  year      = {2024},
  doi       = {10.1609/aaai.v38i1.27766}
}

@article{thuerey2020deep,
  author  = {Thuerey, Nils and Wei{\ss}enow, Konstantin and Prantl, Lukas and Hu, Xiangyu},
  title   = {Deep Learning Methods for {Reynolds}-Averaged {Navier--Stokes} Simulations of Airfoil Flows},
  journal = {AIAA Journal},
  volume  = {58},
  number  = {1},
  pages   = {25--36},
  year    = {2020}
}

@article{geuzaine2009gmsh,
  author  = {Geuzaine, Christophe and Remacle, Jean-Fran{\c{c}}ois},
  title   = {Gmsh: A {3-D} Finite Element Mesh Generator with Built-in Pre- and Post-processing Facilities},
  journal = {Int. J. Numer. Methods Eng.},
  volume  = {79},
  number  = {11},
  pages   = {1309--1331},
  year    = {2009}
}

@article{desbrun2005discrete,
  author  = {Desbrun, Mathieu and Hirani, Anil N. and Leok, Melvin and Marsden, Jerrold E.},
  title   = {Discrete Exterior Calculus},
  journal = {arXiv preprint math/0508341},
  year    = {2005}
}

@article{lim2020hodge,
  author  = {Lim, Lek-Heng},
  title   = {Hodge {Laplacians} on Graphs},
  journal = {SIAM Review},
  volume  = {62},
  number  = {3},
  pages   = {685--715},
  year    = {2020}
}

@article{schaub2020random,
  author  = {Schaub, Michael T. and Benson, Austin R. and Horn, Paul and Lippner, G{\'a}bor and Jadbabaie, Ali},
  title   = {Random Walks on Simplicial Complexes and the Normalized Hodge {1-Laplacian}},
  journal = {SIAM Review},
  volume  = {62},
  number  = {2},
  pages   = {353--391},
  year    = {2020},
  doi     = {10.1137/18m1201019}
}

@article{smirnov2021hodgenet,
  author  = {Smirnov, Dmitriy and Solomon, Justin},
  title   = {{HodgeNet}: Learning Spectral Geometry on Triangle Meshes},
  journal = TOG,
  volume  = {40},
  number  = {4},
  pages   = {1--11},
  year    = {2021}
}

@article{sharp2022diffusionnet,
  author  = {Sharp, Nicholas and Attaiki, Souhaib and Crane, Keenan and Ovsjanikov, Maks},
  title   = {{DiffusionNet}: Discretization Agnostic Learning on Surfaces},
  journal = TOG,
  volume  = {41},
  number  = {3},
  pages   = {1--16},
  year    = {2022}
}

@article{shaffer2026structure,
  author  = {Shaffer, Benjamin and Koohy, Shawn and Kinch, Brooks and Hsieh, M. Ani and Trask, Nathaniel},
  title   = {Structure-Preserving Learning Improves Geometry Generalization in Neural {PDEs}},
  journal = {arXiv preprint arXiv:2602.02788},
  year    = {2026}
}

@article{zheng2026topology,
  author  = {Zheng, Dongzhe and Zhong, Tao and Allen-Blanchette, Christine},
  title   = {Topology-preserving neural operator learning via Hodge decomposition},
  journal = {arXiv:2605.13834},
  year    = {2026}
}

@article{bastian2026topological,
  author  = {Bastian, Lennart and Leventhal, Samuel and Hajij, Mustafa and Birdal, Tolga},
  title   = {Topological Neural Operators},
  journal = {arXiv preprint arXiv:2606.09806},
  year    = {2026}
}

@inproceedings{shazeer2017moe,
  title={Outrageously Large Neural Networks: The Sparsely-Gated Mixture-of-Experts Layer},
  author={Shazeer, Noam and others},
  booktitle={Int. Conf. Learn. Represent.},
  year={2017}
}

@article{fedus2022switch,
  title={Switch Transformers: Scaling to Trillion Parameter Models with Simple and Efficient Sparsity},
  author={Fedus, William and Zoph, Barret and Shazeer, Noam},
  journal={Journal of Machine Learning Research},
  volume={23},
  number={120},
  pages={1--39},
  year={2022}
}

@inproceedings{eliasof2020diffgcn,
  title={{DiffGCN}: Graph Convolutional Networks via Differential Operators and Algebraic Multigrid Pooling},
  author={Eliasof, Moshe and Treister, Eran},
  booktitle={Advances in Neural Information Processing Systems (NeurIPS)},
  year={2020}
}

@inproceedings{tancik2020fourier,
  title     = {Fourier Features Let Networks Learn High Frequency Functions in Low Dimensional Domains},
  author    = {Tancik, Matthew and Srinivasan, Pratul P. and Mildenhall, Ben and Fridovich-Keil, Sara
               and Raghavan, Nithin and Singhal, Utkarsh and Ramamoorthi, Ravi and Barron, Jonathan T.
               and Ng, Ren},
  booktitle = NeurIPS,
  year      = {2020}
}

@inproceedings{smith2019onecycle,
  title     = {Super-Convergence: Very Fast Training of Neural Networks Using Large Learning Rates},
  author    = {Smith, Leslie N. and Topin, Nicholay},
  booktitle = {Artificial Intelligence and Machine Learning for Multi-Domain Operations Applications},
  year      = {2019}
}

@inproceedings{loshchilov2019adamw,
  title     = {Decoupled Weight Decay Regularization},
  author    = {Loshchilov, Ilya and Hutter, Frank},
  booktitle = ICLR,
  year      = {2019}
}

\clearpage

\setcounter{section}{0}
\setcounter{table}{0}
\setcounter{figure}{0}
\setcounter{equation}{0}
\setcounter{proposition}{0}
\setcounter{remark}{0}

\renewcommand{\thesection}{S\arabic{section}}
\renewcommand{\thetable}{S\arabic{table}}
\renewcommand{\thefigure}{S\arabic{figure}}
\renewcommand{\theequation}{S\arabic{equation}}
\renewcommand{\theproposition}{S\arabic{proposition}}
\renewcommand{\theremark}{S\arabic{remark}}

\renewcommand{\theHsection}{S\arabic{section}}
\renewcommand{\theHsubsection}{S\arabic{section}.\arabic{subsection}}
\renewcommand{\theHsubsubsection}{S\arabic{section}.\arabic{subsection}.\arabic{subsubsection}}
\renewcommand{\theHtable}{S\arabic{table}}
\renewcommand{\theHfigure}{S\arabic{figure}}
\renewcommand{\theHequation}{S\arabic{equation}}
\renewcommand{\theHproposition}{S\arabic{proposition}}
\renewcommand{\theHremark}{S\arabic{remark}}


\phantomsection
\pdfbookmark[0]{Supplementary Material}{supplementary}

{\LARGE\bf Supplementary Material\par}

The proposed Hodge-routed mixture of experts uses a controller with vertex, edge, and face updates (VEF). The vertex-only configuration ($V$) is an additional comparison. The supplementary material gives notation, HodgeFormer background, implementation details, training and evaluation settings, model configurations, and results. The controller comparison reports $V$ and VEF separately. The training curves show the vertex-only run. The analyzer and routing measurements use its checkpoint at epoch $380$.


\section{Notation}
\label{sec:notation}

\Cref{tab:notation} defines the main symbols used in the paper and supplementary material. Shared constants and VEF-specific quantities are listed together. Array dimensions do not include the batch dimension. The symbols $x_v,x_e,x_f$ denote input features in \cref{sec:method-features} and normalized features used to compute the mesh operators. The dimensions are given with each equation. Controller stars and analyzer stars are separate operators. Superscripts on decoder tokens indicate block depth. Subscripts on assigned targets identify experts. In the embedding notation, $\delta_s=\operatorname{diag}(d_s^\top d_s)$ with $d_v=d_0$ and $d_e=d_1$, and $A_s$ is the normalized adjacency matrix with its diagonal removed. In the HodgeFormer background (\cref{sec:hodgeformer-preliminaries}), $h$ counts attention heads and $\sigma$ denotes softmax. Elsewhere, these symbols denote the shared embedding and Gaussian bandwidth.

{\small
\setlength{\tabcolsep}{3pt}
\begin{longtable}{@{}ll p{0.19\textwidth}@{\hspace{5pt}}|@{\hspace{5pt}}ll p{0.19\textwidth}@{}}
\caption{\textbf{Notation.} Main symbols grouped by their role. Array dimensions do not include the batch dimension. Uses specific to the HodgeFormer and transformer equations are identified.}
\label{tab:notation}\\
\toprule
symbol & type & meaning & symbol & type & meaning \\
\midrule
\endfirsthead
\multicolumn{6}{@{}l}{\small\emph{\Cref{tab:notation}, continued}}\\
\toprule
symbol & type & meaning & symbol & type & meaning \\
\midrule
\endhead
\bottomrule
\endfoot

\multicolumn{6}{@{}l}{\emph{The discretisation}}\\*
$\M$ & $(V,E,F)$ & the working mesh as a simplicial complex & $e=(i,j)$ & index & an edge with endpoints $i$, $j$ \\
$V,E,F$ & sets & vertices, edges, faces & $i,j,k,m,s$ & indices & elements $i,j$; expert $k$; atom $m$; stream $s$ \\
$\Nv,\Ne,\Nf$ & integers & $|V|,|E|,|F|$; $\Nv\approx4900$ & $c$ & index & output channel in $y_{vc}$ \\
$v$ & index & a vertex &  &  &  \\
\addlinespace
\multicolumn{6}{@{}l}{\emph{Input features}}\\*
$\xv$ & $\R{\Nv\times c_v}$ & standardized vertex inputs & $(c_v,c_e,c_f)$ & $(9,21,13)$ & channel counts \\
$\xe$ & $\R{\Ne\times c_e}$ & edge geometry inputs, not standardized & $\mathrm{pos}$ & $\R{\Nv\times2}$ & standardized position columns of $\xv$ \\
$\xf$ & $\R{\Nf\times c_f}$ & face geometry inputs, not standardized &  &  &  \\
\addlinespace
\multicolumn{6}{@{}l}{\emph{Target and prediction}}\\*
$\ux,\uy$ & fields & velocity components & $\Uinf$ & scalar & freestream speed of a sample \\
$p$ & field & kinematic pressure & $y$ & $\R{\Nv\times4}$ & training target, standardised \\
$\nut$ & field & turbulent kinematic viscosity & $\hat u$ & $\R{\Nv\times4}$ & standardized field prediction \\
\addlinespace
\multicolumn{6}{@{}l}{\emph{Discrete operators}}\\*
$\dzero$ & incidence & $\R{\Ne\times\Nv}$, $(\dzero x)_e=x_j-x_i$ & $\Lzero$ & $\R{\Nv\times\Nv}$ & vertex operator; stars depend on context \\
$\done$ & incidence & $\R{\Nf\times\Ne}$, $\done\dzero=0$ & $Q$ & $\R{\Nv\times\Nv}$ & controller matrix $\dzero^{\!\top}\star_1\dzero$ \\
$\star_0,\star_1,\star_2$ & matrices & stars on $0$-, $1$-, $2$-cochains & $M$ & integer & atom count, $M=9$ \\
$\star_0^{-1},\star_1^{-1}$ & matrices & controller: independent learned operators &  &  &  \\
\addlinespace
\multicolumn{6}{@{}l}{\emph{Neighbourhoods and supports}}\\*
$\mathcal N^{\mathrm{full}}(i)$ & set & stored neighbors: vertex/edge cap $256$, face cap $1$ & support & pairs & the $(i,j)$ with $j\in\mathcal N(i)$ \\
$\mathcal N(i)$ & set & effective neighbors; caps in \cref{tab:arch} & mutual & property & $j\in\mathcal N(i)\Rightarrow i\in\mathcal N(j)$ \\
$n_{vcap},n_{ecap}$ & $64,32$ & VEF vertex and edge caps & $V$, VEF & configurations & vertex-only; vertex, edge, and face updates \\
\addlinespace
\multicolumn{6}{@{}l}{\emph{The controller (\cref{sec:controller})}}\\*
$\ell$ & $\R{\Nv\times K}$ & vertex routing logits & $V^{(i)}$ & array & projected values of head $i$; shape depends on stream \\
$h_v,h_e,h_f$ & arrays & controller streams, each of width $d_c$ & $W_V^{(i)},W_{\text{out}}$ & matrices & per-head value; per-layer output \\
$d_c,L_c,H$ & $64,2,4$ & width, layers, heads & $W_a,W_b$ & matrices & read-out, $d_c\to64\to K$ \\
$d_h$ & $16$ & $d_c/H$, per-head width & $\mathrm{Emb}_s,\mathrm{FFN}_s$ & networks & embedding; feed-forward \\
$A_s$ & matrix & normalized off-diagonal embedding adjacency & $\delta_s,\tilde D_s$ & vector/matrix & incidence degrees; $\tilde D_s=\operatorname{diag}(\delta_s)+I$ \\
\addlinespace
\multicolumn{6}{@{}l}{\emph{The Gaussian star (\cref{sec:controller})}}\\*
$\varphi_i$ & $\R{d_h}$ & kernel feature, $W_\varphi\mathrm{LN}(h)_i$ & $N_c$ & integer & element count in the feature array \\
$W_\varphi$ & $\R{d_h\times d_c}$ & shared kernel projection for elements $i,j$ & $\sigma$ & $>0$ & bandwidth, $\mathrm{softplus}(\gamma)+10^{-8}$ \\
$w_{ij}$ & $(0,1]$ & Gaussian kernel weight & $\beta$ & $(0,\tfrac12)$ & $\tfrac12\operatorname{sigmoid}(\alpha)$ \\
$\Omega$ & $\R{N_c\times N_c}$ & $w_{ij}$ on the support, else $0$ & $\gamma,\alpha$ & scalars & raw parameters; init $0.5413$, $-0.85$ \\
$D,D_i$ & matrix/scalar & $D=\operatorname{diag}(D_i)$; $D_i=\sum_j\Omega_{ij}$ & $A$ & matrix & unweighted vertex adjacency in fixed-band decomposition \\
$\hat S$ & matrix & implemented normalization in \cref{eq:psd-star} &  &  &  \\
\addlinespace
\multicolumn{6}{@{}l}{\emph{Experts, gate and decoder (\cref{sec:experts})}}\\*
$K$ & $8$ & number of experts & $W^k_{\text{in}},W^k_b$ & matrices & expert input; band head \\
$d,d_x$ & $192,88$ & token width; expert width & $\operatorname{proj}_k$ & projection & maps $z_k$ to an $N\times d$ array, with bias \\
$L_x,h_x$ & $2,8$ & blocks; heads per block & $\hat b_k$ & $\R{\Nv\times4}$ & band-head output \\
$h$ & $\R{\Nv\times d}$ & token embedding, shared & $\gate(v)$ & $\simplex$ & gate weights at vertex $v$ \\
$z_k,z^0_k$ & $\R{\Nv\times d_x}$ & output, input of expert $k$ & $\simplex$ & simplex & $K$ nonnegative weights summing to one \\
$n_F$ & $8$ & frequencies; sine/cosine give $2n_F$ channels & $h_{\text{mix}}$ & $\R{\Nv\times d}$ & combined expert features \\
$\sigma_k$ & constants & fixed Fourier scales, $1.0$ to $38.0$ (\cref{tab:arch}) & $E_{\text{field}}$ & $\R{4\times d}$ & decoder field embedding \\
$G_k,B_k$ & $\R{2\times n_F}$ & fixed normal samples; $B_k=\sigma_kG_k$ &  &  &  \\
\addlinespace
\multicolumn{6}{@{}l}{\emph{The analyzer (\cref{sec:analyzer})}}\\*
$w_e$ & $\ge\epsilon_w$ & edge conductance & $\bar y^{(j)}$ & $\R{\Nv\times4}$ & target after $m_j$ steps \\
$\epsilon_w$ & $10^{-4}$ & keeps $w_e$ positive & $m_j,L$ & integers & $m_j$ smoothing steps; $L=8$ levels \\
$\mu_v$ & $>0$ & incident conductance sum; mass uses floor $\epsilon_w$ & $a_i$ & $\R{\Nv\times4}$ & atom $i$; $\sum_i a_i=y$ \\
$\mathrm{MLP}_e$ & $21\to64\to1$ & edge-conductance score & $P,\theta$ & $\R{K\times M}$ & column-normalized assignment and its logits \\
$\tau$ & $\tfrac12$ & forward-Euler step & $t_k$ & $\R{\Nv\times4}$ & band target; $\sum_k t_k=y$ \\
\addlinespace
\multicolumn{6}{@{}l}{\emph{The objective (\cref{sec:method-objective})}}\\*
$\mathcal L$ & scalar & the training objective & $\bar\gate_k$ & scalar & mean routing weight over vertices for expert $k$ \\
$\mathcal L_{\text{vol}},\mathcal L_{\text{surf}}$ & scalars & interior, wall field terms & $p_k$ & $[0,1]$ & assigned-target squared-norm share, denominator floored \\
\addlinespace
\multicolumn{6}{@{}l}{\emph{Metrics}}\\*
$\relL$ & $\ge0$ & mean case/field relative $L_2$, physical units & $\CL$ & $\ge0$ & relative $L_2$ of lift-coefficient vectors \\
$R^2$ & $\le1$ & coefficient of determination & $\rhoL$ & $[-1,1]$ & Spearman correlation of lift \\
Volume, Surf & $\ge0$ & globally normalized MSE, non-wall and wall &  &  &  \\
\addlinespace
\multicolumn{6}{@{}l}{\emph{Generic operators}}\\*
$\mathrm{LN}$ & operator & layer normalisation & $\lambda(\cdot)$ & set & the spectrum of a matrix \\
$\mathrm{Drop}$ & operator & dropout, probability $0.1$ & $\sim$ & relation & similarity of matrices \\
$\mathrm{GELU},\mathrm{MHSA}$ & operators & activation; self-attention & $[\,\cdot\,]$ & operator & concatenation; axis specified locally \\
$\operatorname{diag}(\cdot)$ & matrix & diagonal from a vector & $\odot$ & operator & elementwise product \\
\addlinespace
\multicolumn{6}{@{}l}{\emph{Additional local notation}}\\*
$L_v,L_e,L_f$ & operators & vertex, edge, face operators in \cref{eq:ctrl-layer} & $B_{\mathrm{batch}}$ & integer & batch size \\
$x_v^{\mathrm{raw}}$ & $\R{N\times9}$ & vertex features before standardization & $A_v^{\mathrm{dual}}$ & area & barycentric dual vertex area \\
$\mu_c^{\mathrm{norm}},s_c^{\mathrm{norm}}$ & scalars & per-channel statistics of scaled training targets & $\widetilde y$ & $\R{N\times4}$ & freestream-scaled targets \\
$T$ & $1$ & assignment temperature in \cref{eq:an-assign} & $E_k$ & scalar & $\sum_{v,c}t_{vck}^2$ \\
$\lambda_b,\lambda_{\mathrm{bal}},\lambda_{\mathrm{floor}}$ & scalars & loss weights $0.5$, $0.01$, $1.0$ & $\rho$ & $0.02$ & minimum assigned-target norm share \\
$e_{vc}$ & scalar & normalized field residual in \cref{eq:method-field-loss} & $V_{\mathrm{vol}},V_{\mathrm{surf}}$ & sets & non-wall and wall vertices \\
$t_{vc}^{r}$ & vector & decoder field token at depth $r$ & $m(v)$ & $\R{d}$ & combined features after normalization and dropout \\
$z_i$ & scalar & diagonal-face kernel score in \cref{eq:method-face-star} &  &  &  \\
\end{longtable}
}


\section{HodgeFormer Background}
\label{sec:appendix-preliminaries}

\subsection{HodgeFormer}
\label{sec:hodgeformer-preliminaries}

HodgeFormer~\citep{nousias2026hodgeformer} extends the Transformer layer to operate on vertex, edge, and face features. Following the notation of the original paper, the feature arrays are
\begin{equation}
  x_k\in\mathbb R^{n_k\times d_k},\qquad k\in\{v,e,f\}.
  \label{eq:prelim-hodge-features}
\end{equation}
For a common embedding dimension $d_k=d$ and $h$ attention heads, each head has dimension $d_h=d/h$. Learnable projection matrices define the query, key, and value representations,
\begin{equation}
\begin{aligned}
  Q_k&=x_kW_{Q_k},\\
  K_k&=x_kW_{K_k},\\
  V_k&=x_kW_{V_k},
\end{aligned}
\qquad W_{Q_k},W_{K_k},W_{V_k}\in\mathbb R^{d\times d_h}.
\label{eq:prelim-hodge-projections}
\end{equation}
The original HodgeFormer parameterizes a Hodge Star through row-wise softmax attention,
\begin{equation}
  \star_k(x_k)=\sigma\!\left(\frac{Q_kK_k^T}{\sqrt{d_h}}\right),
  \label{eq:prelim-hodge-star}
\end{equation}
where $\sigma(\cdot)$ denotes the row-wise softmax. Separate linear maps are used for the inverse Hodge stars. The learned operators are the following products of star and signed incidence matrices:
\begin{equation}
\begin{aligned}
  L_v&:=\star_0^{-1}(x_v)\cdot d_0^T\cdot\star_1(x_e)\cdot d_0,\\
  L_e&:=d_0\cdot\star_0^{-1}(x_v)\cdot d_0^T\cdot\star_1(x_e)\\
     &\quad+\star_1^{-1}(x_e)\cdot d_1^T\cdot\star_2(x_f)\cdot d_1,\\
  L_f&:=d_1\cdot\star_1^{-1}(x_e)\cdot d_1^T\cdot\star_2(x_f).
\end{aligned}
\label{eq:prelim-hodge-operators}
\end{equation}
For one head, the attention updates apply each operator to the corresponding projected values,
\begin{equation}
\begin{aligned}
  A_v(x_v,x_e)&=L_v(x_v,x_e)\cdot V_v(x_v),\\
  A_e(x_v,x_e,x_f)&=L_e(x_v,x_e,x_f)\cdot V_e(x_e),\\
  A_f(x_e,x_f)&=L_f(x_e,x_f)\cdot V_f(x_f).
\end{aligned}
\label{eq:prelim-hodge-attention}
\end{equation}
The head outputs are joined along the feature dimension. A linear projection restores the embedding width. These operations define multi-head Hodge attention $A_{H_l}$. The HodgeFormer layer is
\begin{equation}
\begin{split}
  H_l(x_v,x_e,x_f):={}&G_l\!\big(
    A_{H_l}(x_v,x_e,x_f)\\
    &\qquad +(x_v,x_e,x_f)\big),
\end{split}
\label{eq:prelim-hodge-layer}
\end{equation}
where $G_l$ applies separate two-layer feedforward networks to the vertex, edge, and face features. Each network uses a residual connection, which adds its input to its output. These equations describe the original HodgeFormer. The Gaussian-star controller used in this work is specified in \cref{sec:controller}.

\section{Methodology}
\label{sec:appendix-methodology}

This section gives additional details of the controller, expert inputs, transformer blocks, mixture, and field decoder described in \cref{sec:method}. Widths and counts are listed in \cref{tab:arch}.

\subsection{Overall Controller Architecture}
\label{sec:appendix-controller-architecture}

The proposed VEF controller updates vertex, edge, and face features through two PSD-Hodge layers, followed by a vertex readout. The vertex-only configuration is described in \cref{sec:appendix-vef}. The layers act on the vertex, edge, and face feature arrays from \cref{sec:method-features}, each of which has $64$ channels. Each layer has four heads of width $16$ and feedforward width $256$. The projections and stars have separate parameters for each layer and head. These indices are omitted from the equations.

All three feature arrays are normalized before attention. With $x_v=\mathrm{LN}(h_v)$, $x_e=\mathrm{LN}(h_e)$, and $x_f=\mathrm{LN}(h_f)$, one head computes
\begin{equation}
\begin{aligned}
  A_v(x_v,x_e)&=L_v(x_v,x_e)\cdot V_v(x_v),\\
  A_e(x_v,x_e,x_f)&=L_e(x_v,x_e,x_f)\cdot V_e(x_e),\\
  A_f(x_e,x_f)&=L_f(x_e,x_f)\cdot V_f(x_f),
\end{aligned}
\label{eq:method-hodge-heads}
\end{equation}
where $V_s(x_s)=x_sW_{V_s}$ is the learned value projection for stream $s\in\{v,e,f\}$. The operators are those of \cref{eq:ctrl-layer}. 

The four head outputs are joined along the feature dimension. Dropout and a linear projection then return $64$ channels. The combined output for feature array $s$ is $A_{H_l,s}$. The layer updates are
\begin{equation}
\begin{aligned}
  \widetilde h_s&=h_s+\mathrm{Drop}(A_{H_l,s}(x_v,x_e,x_f)),\\
  h_s^{\mathrm{out}}&=\widetilde h_s+
    \mathrm{Drop}(\mathrm{FFN}_s(\mathrm{LN}(\widetilde h_s))),\\
  \mathrm{FFN}_s(x)&=\mathrm{Linear}_{256\to64}
    (\mathrm{ReLU}(\mathrm{Linear}_{64\to256}(x))).
\end{aligned}
\label{eq:method-hodge-block}
\end{equation}

Each feature array has its own normalization layers, projections, and feedforward network. The feedforward updates with residual connections form $G_l$. Dropout has probability $0.1$. The first sublayer applies the mesh operators. The second applies a feedforward network separately to each mesh element. All three attention updates use the features from the start of the layer.

The vertex operator uses edge features through $\star_1$. The edge operator uses vertex and face features through $\star_0^{-1}$ and $\star_2$. The first layer therefore uses face features to update edge features. The second layer can then use those edge features to update vertices. The vertex readout does not directly use the final face update.

After the second PSD-Hodge layer, the controller reads only $h_v$ and computes
\begin{equation}
  \ell=\mathrm{Linear}_{64\to8}
    (\mathrm{GELU}(\mathrm{Linear}_{64\to64}(\mathrm{LN}(h_v))))
    \in\mathbb R^{N\times K},\qquad K=8.
\label{eq:appendix-controller-readout}
\end{equation}
The controller outputs scores before softmax, called logits. The mixture module converts these scores into routing weights using softmax, without temperature scaling. Including the batch dimension, the score array has shape $(B_{\mathrm{batch}},N,K)$.

\subsection{Conditional spectral properties}
\label{sec:appendix-spectral-properties}

The Gaussian star in \cref{sec:controller} is symmetric positive definite (SPD) under the following conditions. The feature array is fixed, and every retained neighbor pair also occurs in the reverse direction. The matrix $\Omega$ of nonnegative kernel weights is then symmetric. With positive row sums $D_i=\sum_j\Omega_{ij}$ and ideal normalization $S=D^{-1/2}\Omega D^{-1/2}$, the matrix $S$ is symmetric and has eigenvalues in $[-1,1]$. Therefore,
\begin{equation}
  \star=(1-\beta)I+\beta S,\qquad
  \lambda(\star)\subseteq[1-2\beta,1],\qquad 0<\beta<\tfrac12,
  \label{eq:appendix-star-spectrum}
\end{equation}
is SPD. This bound uses the ideal normalization. The implementation adds a small constant to the denominator. After all truncation steps, the retained lists must be checked for neighbor pairs missing their reverse. The configured procedure does not ensure that both directions are retained (\cref{sec:appendix-supports}). The identity term alone does not make an asymmetric star symmetric.

If the controller stars $A=\star_0^{-1}$ and $\star_1$, which have separate parameters, are SPD, then $Q=d_0^\top\star_1d_0$ is symmetric positive semidefinite (PSD), and
\begin{equation}
  A^{-1/2}L_vA^{1/2}=A^{1/2}QA^{1/2}
  \label{eq:appendix-vertex-similarity}
\end{equation}
is symmetric PSD. Thus $L_v=AQ$ has real, nonnegative eigenvalues under these assumptions. The notation $\star_0^{-1}$ names an operator learned separately.

The two terms of the VEF edge operator $L_e$ contain stars with separate parameters. Individual SPD stars do not ensure that both terms are symmetric and PSD under the same inner product. The proof for the vertex operator therefore does not prove that the summed edge operator has real, nonnegative eigenvalues. The complete controller block also contains value and output projections, residual connections, and nonlinear transformations. It is not a PSD linear operator.

\subsection{Expert Fourier Bandwidths}
\label{sec:appendix-expert-fourier-bandwidths}

Each expert receives Fourier features of the standardized vertex coordinates $\mathrm{pos}\in\mathbb R^{N\times2}$. For expert $k=1,\ldots,K$, the frequency matrix is $B_k=\sigma_kG_k\in\mathbb R^{2\times n_F}$, where $n_F=8$ and the entries of $G_k$ are sampled independently from a standard normal distribution. The matrices are sampled at initialization and remain fixed during training. The configured scales are

\begin{equation}
(\sigma_1,\ldots,\sigma_K)=(1.0,1.7,2.8,4.8,8.0,13.5,22.6,38.0).
  \label{eq:method-expert-bandwidths}
\end{equation}

The product $\mathrm{pos}\,B_k\in\mathbb R^{N\times n_F}$ gives eight coordinate projections per vertex. Applying sine and cosine elementwise produces the Fourier features
\begin{equation}
  [\sin(2\pi\,\mathrm{pos}\,B_k)\,;\,\cos(2\pi\,\mathrm{pos}\,B_k)]
  \in\mathbb R^{N\times2n_F}.
  \label{eq:appendix-fourier-features}
\end{equation}
The semicolon joins arrays along the feature dimension. Larger $\sigma_k$ increase the sampled frequencies, giving the experts different positional inputs. These scales do not restrict an expert to a particular frequency band. The learned atom assignment determines the auxiliary targets and need not follow the order of the Fourier scales. The experts use the Fourier features as inputs. The analyzer constructs their targets separately, using the smoothing operations and differences between smoothed fields described in \cref{sec:method}.

\subsection{Initialization and parameter counts}
\label{sec:appendix-initialization}

After the model is created, Xavier-uniform initialization is applied to every parameter tensor with more than one dimension, including field embeddings and assignment logits. The fixed Fourier matrices are stored separately from trainable parameters and are unchanged by this step. Evaluation logs report 3,106,490 parameters marked as trainable for the proposed VEF configuration and 3,023,962 for the vertex-only comparison. These totals include unused parameters and analyzer parameters that receive no gradients from the training loss.

\subsection{Analyzer and assignment gradients}
\label{sec:appendix-analyzer-gradients}

The analyzer uses a separate diagonal-star vertex operator to construct nine target components, called atoms, from smoothed reference fields and their differences. An edge-feature network computes the conductances. Before multiplication by the learned assignment matrix, the atoms are treated as constants when computing gradients. The auxiliary band loss and the penalty for assigned-target shares below the threshold therefore update the assignment logits without updating the conductance network. The conductance parameters remain at initialization. The analyzer is used only during training. The auxiliary expert predictions are not inputs to the final decoder.

Each column of the assignment sums to one, so the assigned targets sum to the reference field by construction. This identity does not require orthogonal atoms or a fixed expert order. The final Xavier initialization replaces the assignment values set by the constructor. The Fourier scales remain fixed and do not require the assignment to follow their order.

\section{Supplementary material on the training objective}
\label{sec:appendix-objective}

This section defines the individual terms of \cref{eq:method-loss}.

The field losses compare the final prediction $\hat u$ with the reference $y$. For each case and output channel, we divide the prediction error by the reference field's standard deviation across vertices, with a lower bound of $0.1$ to avoid division by a very small value. We then compute separate mean squared errors on the non-wall vertices $V_{\mathrm{vol}}$ and wall vertices $V_{\mathrm{surf}}$:
\begin{equation}
  e_{vc}=\frac{\hat u_{vc}-y_{vc}}
  {\max(\operatorname{std}_{v}(y_{vc}),0.1)},\qquad
  \mathcal L_{\mathrm{vol}}=\operatorname{mean}_{v\in V_{\mathrm{vol}},c}e_{vc}^2,
  \quad\mathcal L_{\mathrm{surf}}=\operatorname{mean}_{v\in V_{\mathrm{surf}},c}e_{vc}^2.
\label{eq:method-field-loss}
\end{equation}
The two vertex sets are disjoint, so each vertex contributes to one field term. Computing their means separately gives wall errors a contribution independent of the number of interior vertices. The auxiliary term compares each expert output $\hat b_k$ with its assigned four-channel target $t_k$ from \cref{eq:an-assign}.

The routing penalty encourages the mean weight $\bar g_k$ of each expert to approach $1/K$. Individual vertices can still use different expert weights. This penalty concerns how the controller combines expert features.

The last term concerns the sizes of the auxiliary targets. We compute each target's squared norm $E_k=\sum_{v,c}t_{vck}^2$ and its normalized share $p_k=E_k/\max(\sum_{k'}E_{k'},10^{-9})$. An expert contributes zero to this penalty when $p_k\ge\rho$ and a squared penalty $(\rho-p_k)^2$ when $p_k<\rho$. This discourages nearly zero auxiliary targets without requiring all targets to have equal norms. It is a soft penalty and does not enforce a hard lower bound on $p_k$. The targets can contain overlapping spectral components, so these shares do not describe an orthogonal partition of the reference field's squared norm.

The atoms are detached before assignment. The auxiliary term and the last penalty can therefore update the assignment scores without passing gradients through the atoms to the analyzer conductances. The entropy of the normalized target norms is logged for diagnostics and does not contribute to the loss.

\section{Experimental Protocol}
\label{sec:appendix-config}

This section gives the implementation and training settings for the proposed VEF configuration and the vertex-only comparison ($V$). \Cref{tab:training,tab:arch} list the training and architecture settings. Lift is computed using the same contour on the working mesh for every model. The separate contour comparison in \cref{sec:appendix-results} uses a subset of cases and the vertex-only checkpoint.

\begin{table}[!ht]
\centering\footnotesize
\caption{\textbf{Training configuration.}}
\label{tab:training}
\begin{tabularx}{\textwidth}{@{}lX@{}}
\toprule
optimiser & AdamW~\citep{loshchilov2019adamw}, weight decay $10^{-4}$ \\
initial stage & $400$ epochs; per-batch OneCycleLR~\citep{smith2019onecycle}, peak learning rate $10^{-3}$, warm-up fraction $0.3$, division factor $25$, final division factor $10^4$; initial rate $4\times10^{-5}$ and final rate approximately $4\times10^{-9}$ \\
additional VEF training & resume from epoch $400$ for $200$ more epochs; per-batch OneCycleLR, peak learning rate $2\times10^{-4}$, warm-up fraction $0.1$, division factors $25$ and $10^4$ \\
batch & one whole mesh \\
precision & automatic mixed precision with \texttt{bfloat16}; incidence-matrix products and all analyzer operations computed in 32-bit floating point \\
dropout & $0.1$ throughout \\
initialisation & Xavier uniform on every parameter of dimension greater than one \\
checkpoints & every $20$ epochs; VEF: $30$ checkpoints, selected epoch $560$; $V$: $20$ checkpoints, selected epoch $380$ \\
hardware & one NVIDIA RTX~4090 with $24$\,GB memory \\
\bottomrule
\end{tabularx}
\end{table}

\begin{table}[!ht]
\centering\footnotesize
\caption{\textbf{Architecture.} Settings of the proposed VEF configuration and the vertex-only comparison. Dropout is $0.1$ throughout; parameter tensors of dimension greater than one are Xavier-uniform initialized.}
\label{tab:arch}
\begin{tabularx}{\textwidth}{@{}llX@{}}
\toprule
part & symbol & value \\
\midrule
inputs & $(c_v,c_e,c_f)$ & $(9,21,13)$ channels on vertices, edges and faces \\
\midrule
embedding & $d$ & $192$; $W_1:9\to192$, GELU, $W_2:192\to192$ (\cref{eq:fwd-embed}) \\
experts & $K$ & $8$ \\
        & $d_x$ & $88$, with $W^k_{\text{in}}:(d+2n_F)=208\to88$ and $\operatorname{proj}_k:88\to192$ \\
        & $L_x$, $h_x$ & $2$ blocks with normalization before attention and feedforward updates; $8$ heads of dimension $11$; feedforward width $4d_x=352$ \\
        & $n_F$, $\sigma$ & $8$ frequencies giving $16$ sine/cosine channels; $\sigma=(1.0,1.7,2.8,4.8,8.0,13.5,22.6,38.0)$ \\
\midrule
controller & $d_c$, $L_c$ & $64$, $2$ PSD-Hodge layers \\
           & $H$ & $4$ heads, feed-forward $4d_c=256$, read-out $64\to64\to K$ \\
           & $\mathcal N$ & VEF: vertex/edge/face caps $64/32/1$; $V$: vertex/edge caps $64/64$ \\
\midrule
analyzer & hidden & $64$ \\
         & smoothing & $(2,4,6,8,12,16,24,32)$ steps, giving $9$ atoms \\
         & $\star_0$ & the sum of incident edge conductances at each vertex, rather than a separately learned vertex weight; the resulting operator has $\lambda_{\max}\le2$ \\
\midrule
decoder & blocks & $2$ blocks with normalization before attention and feedforward updates; $4$ heads attending across the four output-field tokens at each vertex; feedforward width $768$ \\
        & read-out & $E_{\text{field}}\in\R{4\times192}$ and a shared $\mathrm{MLP}:192\to64\to1$ \\
\midrule
parameters & total & $3{,}106{,}490$ (VEF); $3{,}023{,}962$ ($V$) \\
\bottomrule
\end{tabularx}
\end{table}

\subsection*{Input arrays and normalization}
\label{sec:appendix-input-normalization}

The input arrays have $9$, $21$, and $13$ channels on vertices, edges, and faces. Vertex inputs are the two physical position coordinates, barycentric dual area, two inlet velocity components, signed distance, two supplied normal components, and the wall indicator. All nine columns are standardized with fixed training statistics. The standardized position columns are also used for the expert Fourier features.

Edge inputs concatenate twelve coordinate entries for the edge endpoints and the opposite vertices of the incident triangles, three components of the normalized mean endpoint normal, and six pairwise distances among those four vertices. Missing opposite-vertex coordinates at a boundary are zero-filled. Face inputs concatenate the nine ordered triangle-coordinate entries, three face-normal components, and the triangle area. Geometry is computed with coordinates embedded at $z=0$. Edge and face arrays are supplied without dataset-level standardization. Later embedding and layer-normalization operations are separate transformations. The term raw edge or face features refers to these unstandardized input arrays.

The target scaling is given in \cref{eq:method-target-scaling}: velocity and turbulent kinematic viscosity are divided by the sample's freestream speed, and pressure by its square. A fixed mean and standard deviation are then applied to each output channel. The statistics are computed over vertices from the selected training cases' fixed \code{s0} meshes, with a configured limit of $100$ cases. Each vertex has equal weight. These per-channel statistics are reused for all training and test cases. The additional VEF training uses the saved statistics from the initial stage. Evaluation reverses standardization and freestream scaling for metrics in physical units. The turbulent-viscosity target remains linear, and its division by speed leaves units of length.

\subsection*{Training cases and checkpoint selection}
\label{sec:appendix-selection}

The Hodge-routed configurations train on all $800$ official training cases. Each time a training case is loaded, one of its available remeshes is selected, from a pool of up to eight. Evaluation uses a fixed mesh for each of the $200$ official test cases. The external baselines reserve $80$ of the $800$ training cases for validation and train on the remaining $720$. Their checkpoints are selected using validation errors.

The reported Hodge-routed checkpoints are selected by mean relative $L_2$ error on the $200$ test cases: epoch $560$ among $30$ checkpoints for VEF, and epoch $380$ among $20$ checkpoints for $V$. The reported test errors are therefore also used to choose the checkpoint. Error differences from the external baselines cannot be attributed to architecture alone because the training cases and checkpoint-selection procedures differ. The external all-input configuration has eight channels and omits the wall-indicator channel used by the Hodge-routed configurations (\cref{sec:appendix-transolver-arms}).

\subsection*{Neighbor lists}
\label{sec:appendix-supports}

Neighbor lists are computed once per mesh and remain fixed during training and evaluation. The stored vertex and edge lists contain at most $m=256$ entries and use graph distances up to $R=8$ steps. For a vertex list, ring $r$ contains vertices $r$ steps away in the vertex-adjacency graph. For an edge list, it contains edges $r$ steps away in the edge-adjacency graph. The stored list for element $i$ is $\mathcal N^{\mathrm{full}}(i)$.

Ring $0$ contains the element itself, $\{i\}$. The remaining places are divided among rings $1$ to $R$ in proportion to $\delta^{\,r-1}$, with $\delta=0.55$ and at least one place per ring. The rounded limits are $(116,64,35,19,11,6,3,2)$. These limits and the self entry can exceed the total cap by one. If too many entries are selected, the sampler keeps the first $m$ in order of increasing graph distance. If a ring contains no more neighbors than allowed, all are kept. Otherwise, the allowed number is sampled uniformly at random. Any remaining places are filled with the nearest unselected elements in graph distance. For every selected neighbor pair, its reverse is then added:
\begin{equation}
  \mathcal N^{\mathrm{full}}(i)\;\leftarrow\;
  \mathcal N^{\mathrm{full}}(i)\cup\{\,j: i\in\mathcal N^{\mathrm{full}}(j)\,\},
  \label{eq:support-closure}
\end{equation}
If adding reverse pairs makes a list longer than $m$, it is shortened to $m$ entries. Each face list contains only the face itself, with $m=1$. Under ideal normalization, a Gaussian star using only self neighbors is the identity. The implementation adds a small constant to the denominator. The VEF face star uses the learned positive diagonal weights in \cref{eq:method-face-star}.

The random seed used to sample each mesh's lists is computed from its name. The lists are saved and reused between epochs. After adding reverse pairs as in \cref{eq:support-closure}, the implementation places the element itself first, sorts the other entries by element index, and keeps at most $m$ entries. This truncation can remove one direction of a neighbor pair.

The controller uses the first entries of these stored lists: $64$ vertex entries and $32$ edge entries for VEF, compared with $64$ vertex and $64$ edge entries for $V$. Each face list contains only itself. Unused positions contain $-1$ and are excluded from the operator. Because entries are sorted by element index, the selected entries need not be the nearest neighbors or contain complete graph-distance rings. Selecting the first entries can also remove one direction of a neighbor pair. The procedure therefore does not ensure that every retained pair occurs in both directions. The eigenvalue bounds in \cref{sec:appendix-spectral-properties} require both directions after all these steps.

Saved geometry files are identified by mesh name. Their names do not record the neighbor-list settings. Reusing a saved file does not rebuild its lists with the current settings. When checking or changing neighbor-list settings, the saved lists must also be checked.




\subsection*{Vertex-only training curves}

\begin{figure}[!tb]
  \centering
  \includegraphics[width=\textwidth]{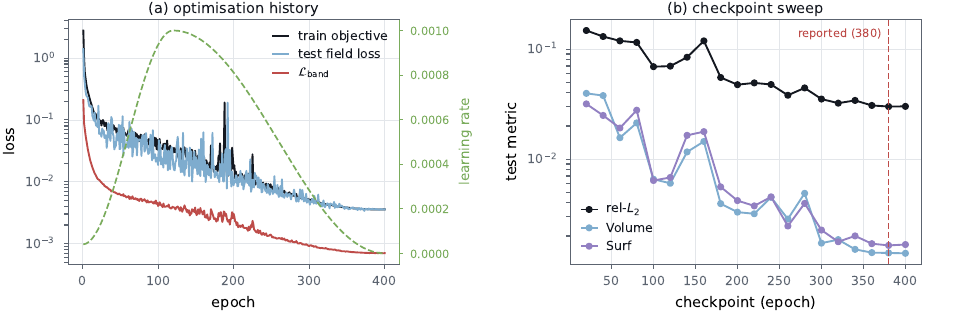}
  \caption{\textbf{Training the vertex-only comparison.} (a) Training objective, test-set field loss, and auxiliary band loss, with the learning rate on the right axis. The training objective includes auxiliary terms absent from the test-set field loss. (b) The $20$-checkpoint test sweep, with epoch $380$ selected by mean relative $L_2$ error. These curves do not describe VEF training.}
  \label{fig:curves}
\end{figure}

\Cref{fig:curves} shows the training curves for the vertex-only comparison. The learning rate reaches $10^{-3}$ at epoch $120$ and decays to approximately $4\times10^{-9}$ by epoch $400$. The training curve includes auxiliary loss terms. The test-set field-loss curve excludes these terms and is computed on the official test loader, although its original logging key is \code{val\_loss}. The auxiliary band loss decreases approximately $300$-fold. Checkpoint evaluation gives vertex-only relative $L_2$ errors of $0.0307$, $0.0300$, and $0.0301$ at epochs $360$, $380$, and $400$, with corresponding Volume errors $0.00141$, $0.00140$, and $0.00139$. The VEF training settings and selected checkpoint are listed separately in \cref{tab:training,tab:vef}.

\section{The Model Family}
\label{sec:appendix-family}

\Cref{tab:family} describes the proposed VEF configuration and the comparison models. Models are grouped by how mesh structure enters field prediction or auxiliary supervision. \Cref{tab:appendix-main} reports the evaluated comparisons.

\paragraph{Mesh-free predictors.}
The internal plain-transformer variants read the nine per-vertex channels of \cref{sec:method-features}. The \emph{plain transformer} uses four pre-normalized softmax transformer layers of width $64$ with global vertex attention, without incidence operators or neighbor supports. The \emph{mesh-windowed} variant restricts attention to the geodesic-ring support, and the \emph{locality-biased} variant adds a learned distance bias to global attention logits. The \emph{wide plain transformer} uses width $192$. The \emph{$+$ lift loss} variant adds a differentiable lift term to its objective.

\paragraph{Frequency mixtures of experts with a feature-space router.}
These models combine transformer experts using per-vertex weights from an MLP on the shared vertex latent. Mesh structure enters the auxiliary target decomposition, without incidence operators in the prediction path. The \emph{fixed-band MoE} uses four experts and a fixed adjacency ladder $(2,8,32)$ with no learned operator: the target is low-passed by powers of the
symmetric normalised adjacency $D^{-1/2}(A+I)D^{-1/2}$, with $A$ the vertex adjacency of the mesh
graph and $D$ its degree matrix, and the four bands are the successive differences of those
low-passes, telescoping to the target as the atoms of \cref{eq:an-atoms} do. This operator is not the
analyzer of \cref{sec:analyzer} with its conductances held fixed: it uses a different normalization and no Euler step size. The two \emph{entropy-balanced} variants use a parameterized Hodge analyzer and maximize the entropy of assigned-target squared-norm shares, with differentiable or detached atom construction, respectively. The configuration named \emph{learned-band MoE} uses eight experts, the diagonal-star analyzer, the learned assignment of \cref{eq:an-assign}, and a one-sided norm-floor penalty. Its atoms are detached, so learning the assignment does not optimize the analyzer conductances (\cref{sec:appendix-analyzer-gradients}). The name denotes learned target assignment rather than prescribed frequency boundaries. The \emph{$+$ wall-surface branch} adds true-surface input channels, a one-dimensional boundary transformer along the wall, and a lift-weighted surface loss. The fixed-band and learned-band configurations also differ in expert count, so their comparison does not isolate the target construction alone.

\paragraph{Hodge-routed mixtures.}
The proposed \emph{Hodge-routed MoE (VEF)} computes gate logits with two PSD-Hodge layers that update vertex, edge, and face streams, followed by a vertex readout. Its prediction experts, target assignment, and field decoder follow the learned-band construction. The \emph{vertex-only} comparison ($V$) uses the vertex Hodge update and retains position-wise edge and face feedforward updates. Its \emph{wide-expert} configuration increases the total model size to $5.68$M, with shared width $256$ and expert width $128$. VEF and $V$ differ in edge support, training duration, and checkpoint count as well as active Hodge updates (\cref{sec:appendix-vef}); their comparison does not isolate cochain degree alone.

\paragraph{Hodge predictors.}
Unlike the controller of \cref{sec:controller}, this line keeps the learned Hodge operator in the
\emph{prediction} path, as in HodgeFormer~\citep{nousias2026hodgeformer}. The \emph{PSD-star
HodgeFormer} learns Gaussian stars, symmetric positive definite under a mutual support (\cref{sec:controller}), and assembles $\Lzero$ from them, with
per-field MLP read-out heads. Adding a cross-field attention decoder, two global blocks and per-case
normalisation gives the \emph{$+$ cross-field decoder} variant, and replacing its loss by the disjoint
volume/surface field terms of \cref{eq:method-field-loss} gives the \emph{$+$ disjoint surface loss} variant, which is the
member with the lowest mean relative $L_2$ error among the three Hodge predictor rows in \cref{tab:appendix-main}. Additional configurations change support symmetry, the vertex star, the bound on $\beta$, the kernel, or layer count.

\paragraph{External baselines.}
Transolver~\citep{wu2024transolver}, PointNet~\citep{qi2017pointnet} and a pointwise MLP are trained
natively on the working meshes and evaluated using the same metric definitions. Training-set allocations and checkpoint selection differ as stated in \cref{sec:setup}. They are named by which channels they receive, where \emph{all inputs} is position, dual area, inlet velocity, signed
distance and wall normals; \emph{pos${+}$inlet} is the minimal arm; and \emph{${+}$SDF${+}$normals}
omits the dual area. These baseline arms have eight channels for all inputs, four for position plus inlet, and seven with SDF and normals; they do not include the ninth wall-indicator channel supplied to our predictor.

\begin{table}[!ht]
\centering\footnotesize
\caption{\textbf{The model family.} Every trained variant in this paper, grouped by how mesh structure
reaches the prediction. \emph{Router} means the operator produces only gate logits; \emph{bands} means
the mesh enters only through the supervisory decomposition of the target; \emph{operator} means the
mesh is in the prediction path; \emph{none} means a point-set model.}
\label{tab:family}
\setlength{\tabcolsep}{4pt}
\begin{tabularx}{\textwidth}{@{}lXr@{}}
\toprule
name used in this paper & what it is & params \\
\midrule
\multicolumn{3}{@{}l}{\emph{mesh-free predictors}, where mesh use is none}\\
Plain transformer                 & $4$-layer pre-norm softmax transformer, width $64$, global attention over all vertices; the Hodge operator deleted & $0.31$M \\
\quad windowed                    & attention restricted to the geodesic-ring support used by the operator & $0.31$M \\
\quad $+$ locality bias           & global attention $+$ a learned soft distance bias on the logits & $0.31$M \\
Wide plain transformer            & the same at width $192$ (FFN $768$, $12$ heads); the width ablation & $2.72$M \\
\quad $+$ lift loss               & $+$ a differentiable $\CL$ contour integral in the objective & $2.72$M \\
\midrule
\multicolumn{3}{@{}l}{\emph{frequency MoE, feature-space (MLP) router}, where mesh use is bands}\\
Fixed-band MoE                    & $4$ band experts, per-vertex gate, fixed adjacency ladder $(2,8,32)$ & $2.75$M \\
\quad entropy-balanced            & parameterized Hodge analyzer with differentiable atoms and an entropy objective & $2.75$M \\
\quad entropy-balanced, detached  & the same with detached atoms & $2.75$M \\
Learned-band MoE                  & $8$ experts, detached analyzer atoms, learned assignment, one-sided norm floor & $2.75$M \\
\quad $+$ wall-surface branch     & $+$ true-surface channels, a 1-D wall transformer and a lift-weighted surface loss & $3.00$M \\
\midrule
\multicolumn{3}{@{}l}{\emph{Hodge-routed MoE}, where mesh use is router}\\
Hodge-routed MoE (VEF)            & proposed model; vertex, edge, and face Hodge updates; vertex readout & $3.11$M \\
Vertex-only controller ($V$)      & vertex Hodge update; position-wise edge and face feedforward updates & $3.02$M \\
\quad wide experts ($V$)          & vertex-only controller; shared width $256$, expert width $128$ & $5.68$M \\
\midrule
\multicolumn{3}{@{}l}{\emph{Hodge predictors}, where mesh use is operator}\\
PSD-star HodgeFormer              & Gaussian stars, $\Lzero$ in the prediction path, per-field heads & $0.45$M \\
\quad $+$ cross-field decoder     & $+$ cross-field attention decoder, $2$ global blocks, per-case normalisation & $0.47$M \\
\quad $+$ disjoint surface loss   & $+$ the disjoint volume/wall MSE of \cref{eq:method-loss} and a wall tag; best of the family & $0.47$M \\
\quad\quad symmetric support      & mutual effective support; conditional vertex spectrum in \cref{sec:appendix-spectral-properties} & $0.47$M \\
\quad\quad no vertex star         & $\star_0^{-1}$ dropped, operator $=\dzero^{\!\top}\star_1\dzero$ & $0.47$M \\
\quad\quad unbound $\beta$        & the $\beta<\tfrac12$ cap released ($\beta$ climbs to $0.86$) & $0.47$M \\
\quad\quad dot-product kernel     & the RBF diffusion kernel replaced by attention on the same support & $0.47$M \\
\quad\quad $1$ / $4$ layers       & the depth sweep of the same operator & $0.35$/$0.72$M \\
\midrule
\multicolumn{3}{@{}l}{\emph{external baselines}, where mesh use is none}\\
Transolver~\citep{wu2024transolver}  & physics-slice attention; arms differ only in input channels & $6.00$M \\
PointNet~\citep{qi2017pointnet}      & shared per-point MLP with permutation-invariant pooling & $0.08$M \\
Pointwise MLP                        & per-vertex MLP, no interaction between vertices, all inputs & $0.02$M \\
\bottomrule
\end{tabularx}
\end{table}

\newpage
\section{Input-Feature Ablation of the Baseline}
\label{sec:appendix-transolver-arms}

The Transolver baseline was retrained on the working mesh at four input arms, so that its row in the
main comparison can be read against the features it was given. The arms differ only in the per-vertex
channels, where \emph{pi} is position and inlet velocity, \emph{pia} adds the dual-cell area, \emph{pisn}
adds the signed distance and the wall normals instead, and \emph{pisna} is all of them, which is the
arm reported in the main paper. All four use the same architecture, the same split and the same
metric code.

\begin{table}[!ht]
\centering\footnotesize
\caption{\textbf{Transolver by input arm.} Metrics on the working meshes for the official $800/200$ split. The training allocation and validation-based checkpoint selection are specified in \cref{sec:appendix-selection}. Best values per metric are in bold.}
\label{tab:transolver-arms}
\setlength{\tabcolsep}{5pt}
\begin{tabular*}{\textwidth}{@{\extracolsep{\fill}}llrcccccc@{}}
\toprule
arm & channels & params & $\relL\downarrow$ & $R^2\uparrow$ & Volume$\downarrow$ & Surf$\downarrow$ & $\CL\downarrow$ & $\rhoL\uparrow$ \\
\midrule
\code{pi}    & pos $+$ inlet                       & $5{,}995{,}332$ & 0.0439 & \best{0.9954} & 0.00270 & 0.00140 & \best{0.0078} & 0.99988 \\
\code{pia}   & $+$ area                            & $5{,}995{,}844$ & 0.0592 & 0.9922 & 0.00418 & 0.00282 & 0.0115 & 0.99980 \\
\code{pisn}  & $+$ SDF $+$ normals                 & $5{,}996{,}868$ & 0.0440 & 0.9952 & 0.00286 & \best{0.00137} & 0.0098 & 0.99977 \\
\code{pisna} & $+$ SDF $+$ normals $+$ area        & $5{,}997{,}380$ & \best{0.0436} & 0.9953 & \best{0.00265} & 0.00143 & 0.0103 & \best{0.99989} \\
\bottomrule
\end{tabular*}
\end{table}

The spread across arms on the field metrics is $0.0436$ to $0.0592$; the highest relative $L_2$ is
\code{pia}, which adds the dual area without the signed distance and the wall normals. The all-input arm has the lowest relative $L_2$ and Volume errors. The position-plus-inlet arm has the highest $R^2$ and lowest lift error, and the SDF-plus-normals arm has the lowest Surf error. The four arms contain $4$, $5$, $7$, and $8$ channels, respectively; none includes the wall-indicator input used by our predictor.

\section{Complete Results}
\label{sec:appendix-results}

\ifdefined\arxivqualinresults\else
\Cref{fig:qual} shows one test case from the vertex-only configuration at epoch $380$, with the reference, prediction, and absolute error for each field. Its per-field relative $L_2$ errors are $0.0047$, $0.0260$, $0.0281$, and $0.0676$, compared with test-set means of $0.0056$, $0.0259$, $0.0255$, and $0.0629$. These results concern the vertex-only configuration rather than VEF. Visible errors occur near the airfoil and along the turbulent-viscosity wake.
\begin{figure}[!ht]
  \centering
  \begin{tikzpicture}
    \node[anchor=south west,inner sep=0pt] (qual)
      {\includegraphics[width=0.92\textwidth]{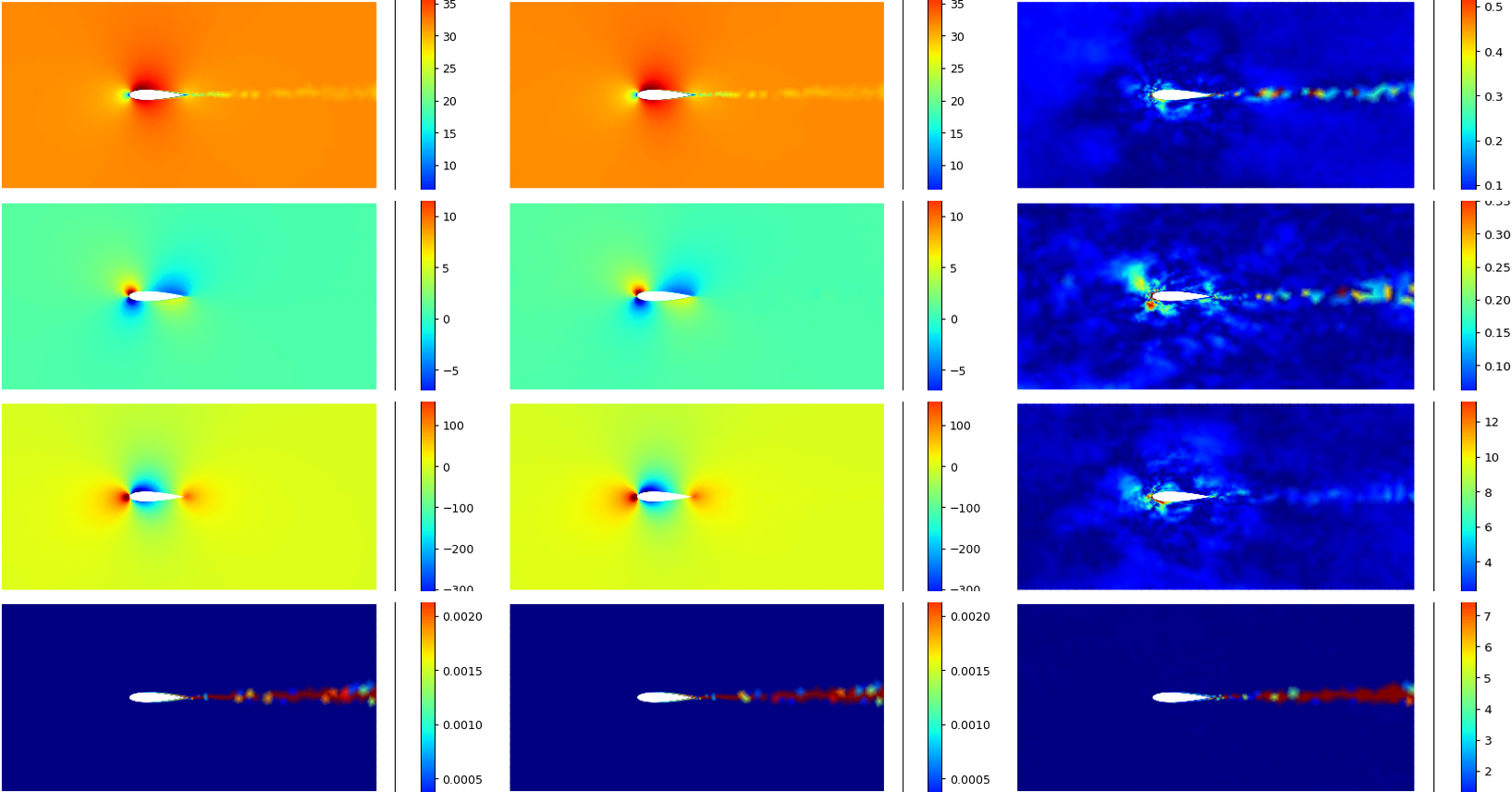}};
    \begin{scope}[x={(qual.south east)},y={(qual.north west)}]
      \node[anchor=south,font=\small] at (0.1248,1) {Reference};
      \node[anchor=south,font=\small] at (0.4606,1) {Prediction};
      \node[anchor=south,font=\small] at (0.8040,1) {Absolute error};
      \node[anchor=south,font=\small,rotate=90] at (0,0.8803) {$\ux$};
      \node[anchor=south,font=\small,rotate=90] at (0,0.6262) {$\uy$};
      \node[anchor=south,font=\small,rotate=90] at (0,0.3732) {$p$};
      \node[anchor=south,font=\small,rotate=90] at (0,0.1197) {$\nut$};
    \end{scope}
  \end{tikzpicture}
  \caption{\textbf{One test case from the vertex-only configuration (epoch $380$).} Columns are the reference,
  the prediction and the absolute error; rows are $\ux$, $\uy$, $p$ and $\nut$. All panels are on the
  model's own working mesh in physical units, with the reference and the prediction sharing each
  field's colour scale. The per-field relative $L_2$ for this case is $0.0047$, $0.0260$, $0.0281$ and
  $0.0676$. Test-set means are listed in \cref{tab:appendix-perfield}; the panel illustrates one case rather than a test-set average.}
  \label{fig:qual}
\end{figure}

\fi

\Cref{tab:appendix-main} reports the proposed VEF configuration and the evaluated comparison models described in \cref{tab:family}. \Cref{tab:appendix-perfield} gives the per-field
breakdown for selected models, \cref{tab:perfield-volsurf} decomposes the two AirFRANS-native
metrics per field, \cref{tab:sweep} reports the checkpoint sweep, and \cref{fig:qual} shows one test
case from the vertex-only checkpoint at epoch $380$. This qualitative example and the diagnostics in \cref{sec:appendix-router} describe the vertex-only comparison.

The VEF results correspond to case \code{v57\_gated\_hodge\_vef} at epoch $560$; the vertex-only results correspond to \code{v57\_gated\_hodge} at epoch $380$. Their field and Volume/Surf values were checked against the saved \code{test\_metrics.npz} files, and their checkpoint counts against the sweep summaries. Complete-comparison rows report all metrics at the checkpoint selected by mean relative $L_2$. The sweep table separately lists the minimum of each stated metric and its epoch.

\begin{table}[!ht]
\centering\footnotesize
\caption{\textbf{Complete comparison.} Official AirFRANS $800/200$ split, evaluation on the
${\approx}4.9$k-vertex mesh, the same test meshes and evaluation metrics for every row. Objectives differ across loss ablations; training and selection protocols are specified in \cref{sec:appendix-selection}. Models are described
in \cref{sec:appendix-family}. Best per column in \textbf{bold}.}
\label{tab:appendix-main}
\setlength{\tabcolsep}{4pt}
\begin{tabular}{@{}llrccccc@{}}
\toprule
model & mesh use & params & $\relL\downarrow$ & $R^2\uparrow$ & Volume$\downarrow$ & Surf$\downarrow$ & $\CL\downarrow$ \\
\midrule
Hodge-routed MoE (VEF)                 & router & $3.11$M & 0.0359 & 0.9978 & 0.00182 & 0.00180 & 0.0277 \\
Hodge-routed MoE ($V$)                 & router & $3.02$M & \best{0.0300} & \best{0.9983} & \best{0.00140} & 0.00164 & 0.0225 \\
\quad wide experts ($V$)              & router & $5.68$M & 0.0341 & 0.9979 & 0.00163 & 0.00219 & 0.0286 \\
Learned-band MoE                       & bands  & $2.75$M & 0.0345 & 0.9979 & 0.00165 & 0.00192 & 0.0265 \\
Fixed-band MoE                         & bands  & $2.75$M & 0.0351 & 0.9978 & 0.00178 & 0.00184 & 0.0280 \\
\quad $+$ wall-surface branch          & bands  & $3.00$M & 0.0378 & 0.9973 & 0.00196 & 0.00227 & 0.0247 \\
\quad entropy-balanced, detached       & bands  & $2.75$M & 0.0402 & 0.9973 & 0.00206 & 0.00214 & 0.0270 \\
\quad entropy-balanced                 & bands  & $2.75$M & 0.0412 & 0.9972 & 0.00198 & 0.00233 & 0.0322 \\
\midrule
Transolver, all inputs~\citep{wu2024transolver} & none & $6.0$M & 0.0436 & 0.9953 & 0.00265 & 0.00143 & 0.0103 \\
Transolver, pos${+}$inlet              & none   & $6.00$M & 0.0439 & 0.9954 & 0.00270 & 0.00140 & \best{0.0078} \\
Transolver, ${+}$SDF${+}$normals       & none   & $6.00$M & 0.0440 & 0.9952 & 0.00286 & \best{0.00137} & 0.0098 \\
\midrule
Wide plain transformer                 & none   & $2.72$M & 0.0446 & 0.9960 & 0.00265 & 0.00192 & 0.0245 \\
\quad $+$ lift loss                    & none   & $2.72$M & 0.0458 & 0.9960 & 0.00248 & 0.00296 & 0.0378 \\
Plain transformer                      & none   & $0.31$M & 0.0544 & 0.9943 & 0.00374 & 0.00300 & 0.0173 \\
PointNet, all inputs~\citep{qi2017pointnet} & none & $0.08$M & 0.0689 & 0.9904 & 0.00451 & 0.00440 & 0.0149 \\
Pointwise MLP, all inputs                      & none     & $0.02$M & 0.1743 & 0.9404 & 0.03576 & 0.03969 & 0.0860 \\
\midrule
PSD-star HodgeFormer, $+$ surface loss & operator & $0.47$M & 0.0800 & 0.9885 & 0.01028 & 0.01048 & 0.0681 \\
PSD-star HodgeFormer, $+$ decoder      & operator & $0.47$M & 0.0822 & 0.9884 & 0.00803 & 0.01205 & 0.0567 \\
PSD-star HodgeFormer                           & operator & $0.45$M & 0.1154 & 0.9769 & 0.01446 & 0.03174 & 0.1747 \\
\bottomrule
\end{tabular}
\end{table}

\begin{table}[!ht]
\centering\footnotesize
\caption{\textbf{Per-field relative $L_2$ for selected models.} Errors for the proposed VEF configuration, the vertex-only configuration ($V$), and comparison models. Metrics use the selected checkpoint of each run. Best values per field are in bold.}
\label{tab:appendix-perfield}
\begin{tabular}{@{}lcccc@{}}
\toprule
model & $\ux$ & $\uy$ & $p$ & $\nut$ \\
\midrule
Hodge-routed MoE (VEF)             & 0.0067 & 0.0348 & 0.0380 & 0.0642 \\
Hodge-routed MoE ($V$)             & \best{0.0056} & 0.0259 & \best{0.0255} & \best{0.0629} \\
\quad wide experts ($V$)          & 0.0072 & 0.0319 & 0.0337 & 0.0638 \\
Learned-band MoE                   & 0.0068 & 0.0288 & 0.0365 & 0.0658 \\
Fixed-band MoE                     & 0.0068 & 0.0312 & 0.0346 & 0.0679 \\
\quad $+$ wall-surface branch      & 0.0064 & 0.0293 & 0.0342 & 0.0812 \\
Transolver, all inputs             & 0.0067 & \best{0.0220} & 0.0298 & 0.1158 \\
Wide plain transformer             & 0.0059 & 0.0330 & 0.0326 & 0.1068 \\
Plain transformer                  & 0.0067 & 0.0554 & 0.0349 & 0.1205 \\
PSD-star HodgeFormer, best         & 0.0149 & 0.0650 & 0.0852 & 0.1548 \\
PointNet, all inputs               & 0.0125 & 0.0540 & 0.0655 & 0.1435 \\
Pointwise MLP, all inputs          & 0.0518 & 0.1865 & 0.1765 & 0.2825 \\
\bottomrule
\end{tabular}
\end{table}

\begin{table}[!ht]
\centering\footnotesize
\caption{\textbf{Per-field Volume and Surf.} AirFRANS-native normalized mean squared errors on non-wall and wall vertices, respectively, at the checkpoint selected by mean relative $L_2$ for each run. VEF and the vertex-only comparison ($V$) are listed separately.}
\label{tab:perfield-volsurf}
\setlength{\tabcolsep}{5pt}
\begin{tabular}{@{}lrrrrcrrrr@{}}
\toprule
& \multicolumn{4}{c}{Volume (interior)} & & \multicolumn{4}{c}{Surf (wall)} \\
\cmidrule(lr){2-5}\cmidrule(lr){7-10}
model & $\ux$ & $\uy$ & $p$ & $\nut$ & & $\ux$ & $\uy$ & $p$ & $\nut$ \\
\midrule
Hodge-routed MoE (VEF)      & 0.00019 & 0.00096 & 0.00077 & 0.00534 & & 0.00079 & 0.00197 & 0.00272 & 0.00173 \\
Hodge-routed MoE ($V$)      & 0.00011 & 0.00052 & 0.00031 & 0.00466 & & 0.00073 & 0.00174 & 0.00248 & 0.00160 \\
Learned-band MoE              & 0.00019 & 0.00063 & 0.00072 & 0.00506 & & 0.00087 & 0.00195 & 0.00307 & 0.00181 \\
Fixed-band MoE                & 0.00019 & 0.00086 & 0.00069 & 0.00539 & & 0.00090 & 0.00198 & 0.00296 & 0.00152 \\
Wide plain transformer        & 0.00017 & 0.00109 & 0.00074 & 0.00859 & & 0.00062 & 0.00191 & 0.00313 & 0.00201 \\
Plain transformer             & 0.00019 & 0.00279 & 0.00060 & 0.01140 & & 0.00088 & 0.00403 & 0.00431 & 0.00278 \\
PSD-star HodgeFormer, best    & 0.00114 & 0.00476 & 0.00420 & 0.03103 & & 0.00406 & 0.00609 & 0.01801 & 0.01378 \\
\bottomrule
\end{tabular}
\end{table}

\begin{table}[!ht]
\centering\footnotesize
\caption{\textbf{Checkpoint sweep.} Separate minimum mean relative $L_2$ and Volume values with their epochs.
$^{\dagger}$\,the VEF
configuration was trained for $400$ epochs plus a $200$-epoch continuation with peak learning rate $2\times10^{-4}$
and swept at $30$ checkpoints rather than $20$.}
\label{tab:sweep}
\begin{tabular}{@{}lrrr@{}}
\toprule
model & checkpoints & best $\relL$ (epoch) & best Volume (epoch) \\
\midrule
Hodge-routed MoE (VEF)$^{\dagger}$ & 30 & 0.0359 (560) & 0.00182 (560) \\
Hodge-routed MoE ($V$)             & 20 & 0.0300 (380) & 0.00139 (400) \\
\quad wide experts ($V$)          & 20 & 0.0341 (400) & 0.00163 (400) \\
Learned-band MoE                   & 20 & 0.0345 (380) & 0.00165 (380) \\
Fixed-band MoE                     & 20 & 0.0351 (400) & 0.00178 (380) \\
Plain transformer                  & 20 & 0.0544 (320) & 0.00339 (400) \\
PSD-star HodgeFormer, best         & 20 & 0.0800 (320) & 0.00651 (400) \\
\bottomrule
\end{tabular}
\end{table}

For VEF, both minimum mean relative $L_2$ and minimum Volume occur at epoch $560$. For the vertex-only comparison, they occur at epochs $380$ and $400$, respectively. The vertex-only Volume value in the complete comparison is $0.00140$ at the relative-$L_2$-selected checkpoint, rather than the separate minimum $0.00139$.

The VEF lift error is $\CL=0.0277$ over $200$ test cases. The vertex-only checkpoint gives $\CL=0.0225$ and $\rhoL=0.99988$. A separate $10$-case contour comparison using the vertex-only checkpoint gives relative lift errors of $0.0220$ for the band and $0.0205$ for the surface loop. These subset values do not establish a full-test-set comparison of the contour conventions or a VEF contour comparison.

\section{Comparison of Controller Configurations}
\label{sec:appendix-vef}

The proposed VEF controller applies $\Lzero$ to vertices, $L_1=L_1^{\text{down}}+L_1^{\text{up}}$ to edges, and $L_2^{\text{down}}$ to faces, using the incidence composition in \cref{eq:ctrl-layer}. The vertex-only configuration ($V$) applies the vertex Hodge attention update without edge or face Hodge attention updates. Its edge and face streams retain position-wise residual feedforward updates; edge features therefore continue to enter the vertex star. Both configurations produce routing logits from the final vertex stream.

The VEF run uses an edge support cap of $32$, compared with $64$ for $V$. It also includes a $200$-epoch continuation and evaluates $30$ checkpoints rather than $20$. The comparison therefore does not isolate cochain degree alone. \Cref{tab:vef} reports both selected runs.

Face supports contain one entry. Under ideal normalization, the self-only Gaussian star is the identity; the implemented denominator offset gives an approximately identity operator. VEF uses the learned positive diagonal face star of \cref{eq:method-face-star}, with $z_i=\lVert\varphi_i\rVert^2/(d_h\sigma^2)$. This scaling is not constrained to equal a geometric face-area star. Incidence applications flatten head and channel axes before sparse multiplication, avoiding a separate sparse call for each head.

\begin{table}[!ht]
\centering\footnotesize
\caption{\textbf{VEF and vertex-only ($V$) controllers.} VEF is the proposed configuration. Training duration, support caps, and checkpoint counts differ between the runs. All errors are evaluated at the checkpoint selected by mean relative $L_2$ over $200$ test cases. The final column gives VEF relative to $V$; percentage changes use unrounded metrics. Bold values are the better of the two configurations for each metric.}
\label{tab:vef}
\begin{tabular}{@{}lccc@{}}
\toprule
& VEF (proposed) & $V$ (vertex-only) & VEF relative to $V$ \\
\midrule
operators           & $\Lzero$, $L_1^{\text{down}}\!+\!L_1^{\text{up}}$, $L_2^{\text{down}}$ & $\Lzero$ & \\
supports            & vertex/edge/face $64/32/1$ & vertex/edge $64/64$ & \\
params              & $3{,}106{,}490$ & $3{,}023{,}962$ & $+82{,}528$ ($+2.7\%$) \\
epochs / checkpoints& $400+200$ / $30$ & $400$ / $20$ & \\
selected checkpoint & $560$ & $380$ & \\
\midrule
$\relL\downarrow$   & 0.0359  & \best{0.0300}  & $+19.7\%$ \\
$R^2\uparrow$       & 0.9978  & \best{0.9983}  & \\
Volume$\downarrow$  & 0.00182 & \best{0.00140} & $+29.7\%$ \\
Surf$\downarrow$    & 0.00180 & \best{0.00164} & $+10.1\%$ \\
$\CL$ (band)$\downarrow$    & 0.0277 & \best{0.0225} & $+23.5\%$ \\
\midrule
$\ux$   & 0.0067 & \best{0.0056} & $+18.5\%$ \\
$\uy$   & 0.0348 & \best{0.0259} & $+34.1\%$ \\
$p$     & 0.0380 & \best{0.0255} & $+48.9\%$ \\
$\nut$  & 0.0642 & \best{0.0629} & $+2.0\%$ \\
\bottomrule
\end{tabular}
\end{table}

The selected VEF run has higher errors than the selected vertex-only run on the metrics in \cref{tab:vef}. Differences in support, training duration, and checkpoint count, together with one run per configuration, limit attribution of the change to the active Hodge updates. Among the configurations in \cref{tab:appendix-main}, the vertex-only Hodge-routed MoE has the lowest mean relative $L_2$ error; other models have lower surface and lift errors.

\section{Diagnostics of the Vertex-only Configuration}
\label{sec:appendix-router}

This section retains the analyzer, assignment, and gate diagnostics for the vertex-only configuration ($V$) at epoch $380$. These figures and measurements do not describe the proposed VEF checkpoint at epoch $560$. The analyzer's diagonal stars are
\begin{equation}
  \star_1=\operatorname{diag}(w_e),\quad w_e=\mathrm{softplus}(\mathrm{MLP}_e(\xe))+\epsilon_w;
  \qquad
  \star_0=\operatorname{diag}(\mu_v),\quad \mu_v=\sum_{e\ni v}w_e ,
\end{equation}
where $w_e$, the edge conductance, is the analyzer's only parameterised content. The atoms are the
telescoping differences of the forward-Euler smoothing ladder $\bar y^{(j)}=(I-\tau\Lzero)^{m_j}y$,
\begin{equation}
  a_0=\bar y^{(L-1)},\qquad
  a_i=\bar y^{(L-1-i)}-\bar y^{(L-i)}\quad(1\le i\le L-1),\qquad
  a_{M-1}=y-\bar y^{(0)} ,
\end{equation}
ordered from the most-smoothed component ($a_0$) to the residual after the fewest smoothing steps ($a_{M-1}$), and the assignment $P$ groups them into per-expert
band targets $t_k$,
\begin{equation}
  P_{km}=\frac{\exp(\theta_{km}/T)}{\sum_{k'}\exp(\theta_{k'm}/T)},
  \qquad \sum_kP_{km}=1\ \ \forall m,
  \qquad t_k=\sum_{m=0}^{M-1}P_{km}\,a_m .
\end{equation}

The analyzer atoms are detached before assignment, so its conductance network receives no gradient from \cref{eq:method-loss} and remains at initialization (\cref{sec:appendix-analyzer-gradients}). Conductance is a function of edge geometry. Its contribution to smoothing is normalized by the incident degree: the weight of neighbor $j$ at vertex $i$ is proportional to $w_{ij}/\mu_i$. Larger unnormalized conductance alone does not establish faster removal of local field structure. Labels referring to a learned analyzer in the retained image panels identify the parameterized module; its conductances were not optimized by this objective.

\begin{figure}[!ht]
  \centering
  \includegraphics[width=\textwidth]{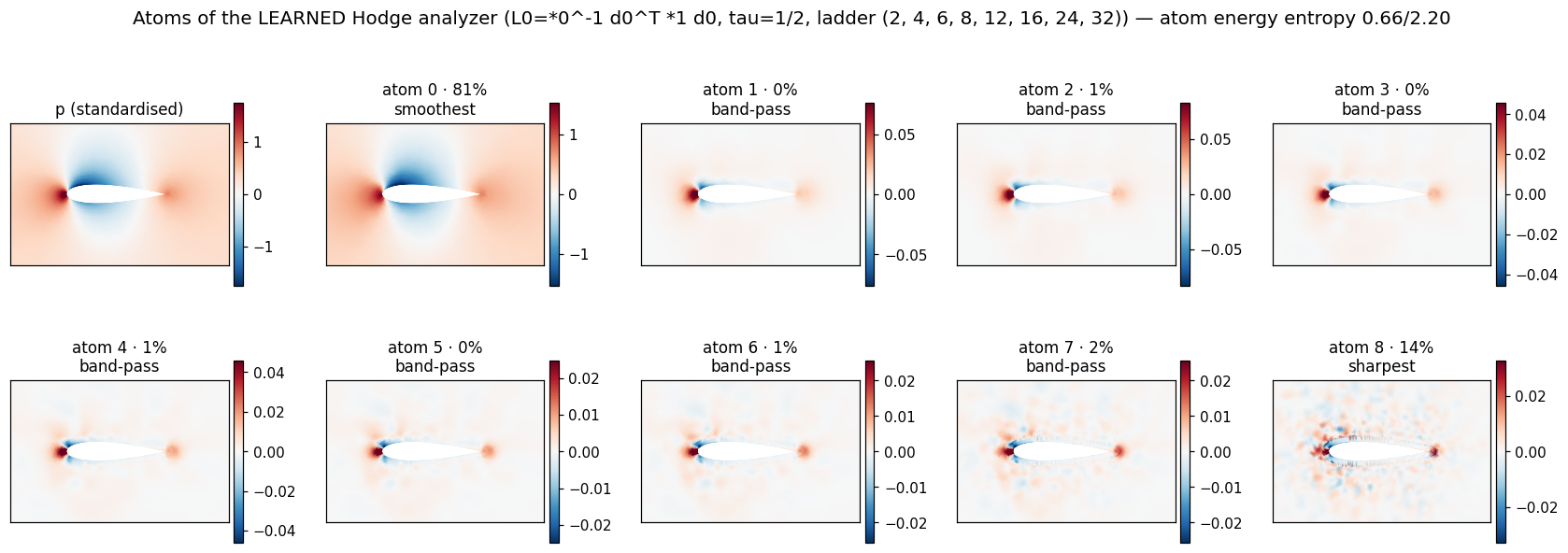}
  \caption{\textbf{Vertex-only analyzer atoms at epoch $380$.} Nine telescoping atoms for the illustrated pressure field. Atom $0$ is the most-smoothed component; atoms $1$--$7$ are differences between smoothing levels, and atom $8$ is the residual after the fewest smoothing steps. The first and last components account for approximately $81\%$ and $14\%$ of the summed squared atom norms. Their sum reconstructs the field algebraically, subject to floating-point rounding (\cref{eq:an-atoms}). The conductance network remains at initialization.}
  \label{fig:atoms}
\end{figure}

\begin{figure}[!ht]
  \centering
  \includegraphics[width=\textwidth]{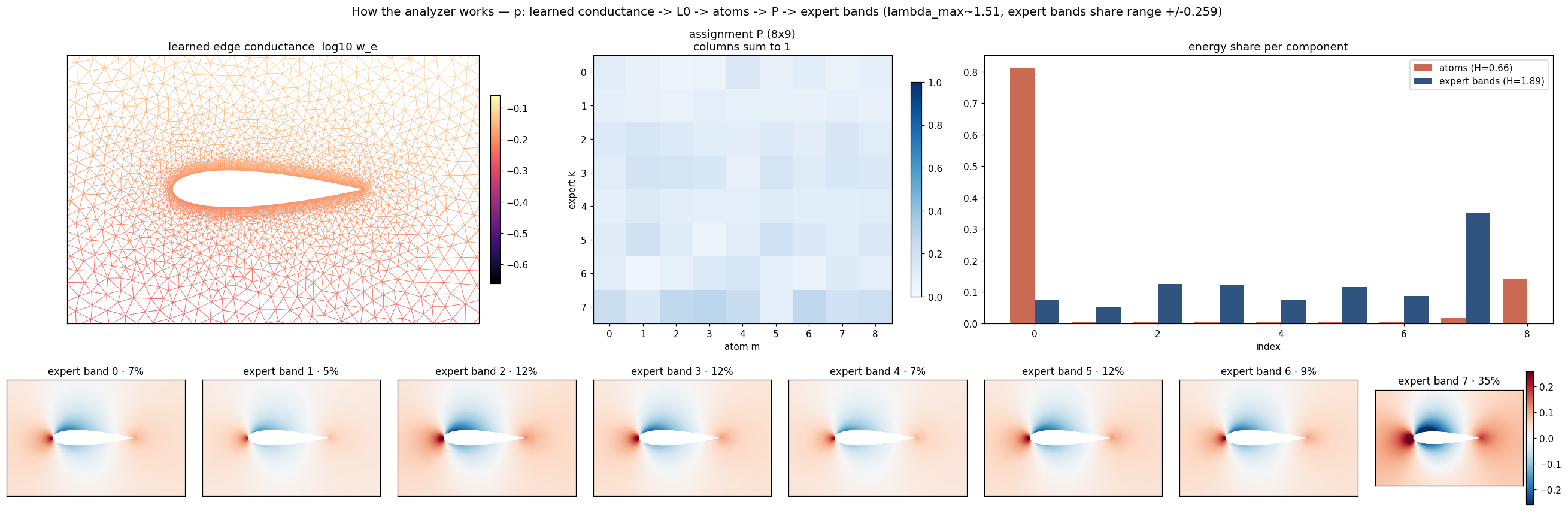}
  \caption{\textbf{Vertex-only analyzer and assignment at epoch $380$.} Edge conductance $w_e$ from the initialized conductance network; the learned $8\times9$ assignment $P$ (\cref{eq:an-assign}); the eight assigned targets on a shared symmetric colour range; and normalized squared component norms before and after assignment. Global Xavier initialization overwrites the constructor's preferred assignment logits.}
  \label{fig:operator}
\end{figure}

\begin{figure}[!ht]
  \centering
  \includegraphics[width=\textwidth]{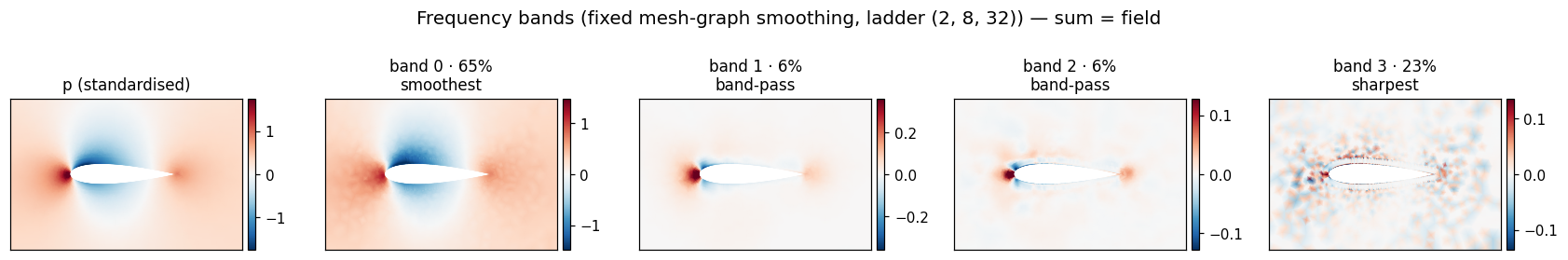}
  \caption{\textbf{The fixed reference decomposition} of the pressure field with the ladder $(2,8,32)$
  on the unweighted adjacency, for comparison with the vertex-only analyzer example in \cref{fig:atoms}. This decomposition is independent of a trained controller checkpoint. The normalized squared component norms for this illustrated field are
  $[0.65,0.06,0.06,0.23]$; the components overlap.}
  \label{fig:bands}
\end{figure}

\Cref{fig:atoms,fig:operator} illustrate the pressure decomposition and assignment at the epoch-$380$ checkpoint. The plotted atom shares are normalized squared norms, $\lVert a_m\rVert^2/\sum_{m'}\lVert a_{m'}\rVert^2$, for the illustrated field. The rounded shares are $[0.812,0.003,0.006,0.003,0.006,0.003,0.006,0.019,0.143]$. Atoms overlap, so these values are not an orthogonal partition of the target's energy. Approximately $81\%$ of the summed squared atom norms is in the coarsest atom and $14\%$ in the finest residual. These case-level diagnostics do not establish corresponding test-set means.

The energy-floor penalty acts on assigned targets $t_k$, not raw atoms $a_m$. Its activation therefore depends on the assigned-target norm shares. The vertex-only training log records a zero floor penalty at epoch $380$; this value does not establish inactivity throughout training. The constructor's preferred atom-to-expert grouping is overwritten by global Xavier initialization, so no prescribed ordering is imposed on the learned assignment.

\begin{figure}[!ht]
  \centering
  \includegraphics[width=\textwidth]{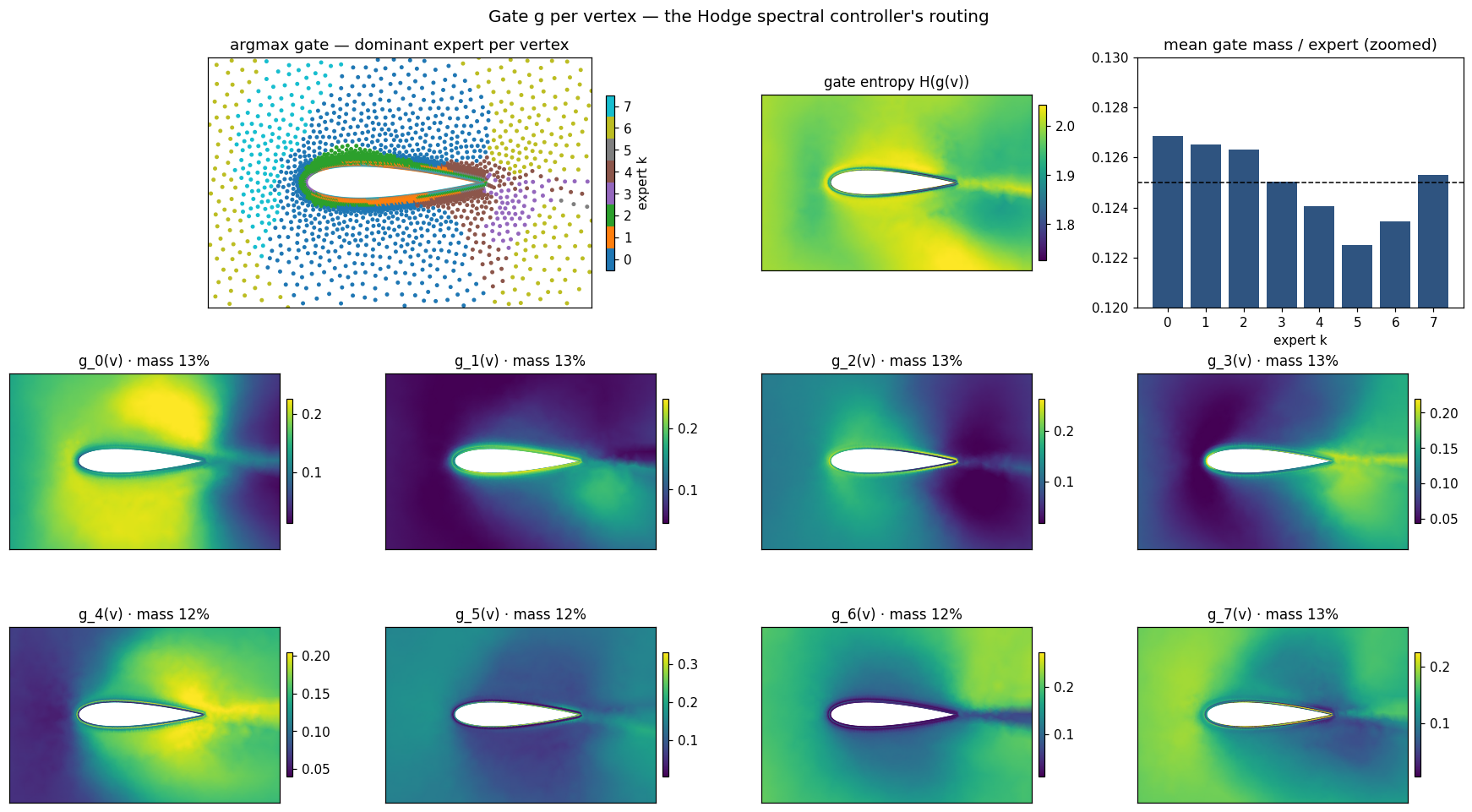}
  \caption{\textbf{Vertex-only gate $\gate(v)$ at epoch $380$.} Top: dominant expert at each vertex, per-vertex gate entropy, and mean routing weight over vertices per expert against the uniform $1/K$ line (vertical axis restricted to $0.120$--$0.130$). Bottom: the eight weight fields $\gate_k(v)$. The dominant-expert panel displays an argmax of soft weights; all experts are evaluated. These panels describe routing for the displayed mesh rather than averages over the test set.}
  \label{fig:gate}
\end{figure}

\end{document}